\RequirePackage{fix-cm}
\documentclass[twocolumn]{svjour3}          % twocolumn
\smartqed  % flush right qed marks, e.g. at end of proof
\usepackage{graphicx}

\usepackage{framed,multirow}
\usepackage{amssymb}
\usepackage{latexsym}

\usepackage{url}
\usepackage{xcolor}
\usepackage{colortbl}
\newcommand{\btxt}[1]{\textcolor{black}{#1}}%blue
\newcommand{\rtxt}[1]{\textcolor{black}{#1}}%red
\definecolor{newcolor}{rgb}{.8,.349,.1}
\usepackage{subcaption}
\begin{document}
	
	\title{Next-best-view Regression using a 3D Convolutional Neural Network\thanks{This work was partially supported by CONACYT-c\'atedra 1507 project.}
	}
	
	
	\author{J. Irving Vasquez-Gomez         \and
		David Troncoso \and Israel Becerra \and Enrique Sucar \and Rafael Murrieta-Cid%etc.
	}
	
	
	\institute{J. Irving Vasquez-Gomez \at
		Consejo Nacional de Ciencia y Tecnolog\'ia (CONACYT) - Insituto Polit\'ecnico Nacional \\
		%\\
		\email{jvasquezg@ipn.mx}           %  \\
		\and
		David Troncoso \at
		Consejo Nacional de Ciencia y Tecnolog\'ia (CONACYT) - Escuela Superior de C\'omputo (ESCOM), Instituto Polit\'ecnico Nacional (IPN), M\'exico City, M\'exico.
		\and
		Israel Becerra \at
		Consejo Nacional de Ciencia y Tecnolog\'ia (CONACYT) - Centro de Investigaci\'on en Matem\'aticas CIMAT, Guanajuato, M\'exico
		\and
		Enrique Sucar \at
		Instituto Nacional de Astrof\'isica \'Optica y Electr\'onica (INAOE), Puebla, M\'exico.
		\and 
		Rafael Murrieta-Cid \at
		Centro de Investigaci\'on en Matem\'aticas CIMAT, Guanajuato, M\'exico
	}
	
	\date{Received: date / Accepted: date}

	\maketitle
	
	\begin{abstract}
		Automated three-dimensional (3D) object reconstruction is the task of building a geometric representation of a physical object by means of sensing its surface. Even though new single view reconstruction techniques can predict the surface, they lead to incomplete models, specially, for non commons objects such as antique objects or art sculptures. Therefore, to achieve the task's goals, it is essential to automatically determine the locations where the sensor will be placed so that the surface will be completely observed. This problem is known as the next-best-view problem. In this paper, we propose a data-driven approach to address the problem. The proposed approach trains a 3D convolutional neural network (3D CNN) with previous reconstructions in order to regress the \btxt{position of the} next-best-view. To the best of our knowledge, this is one of the first works that directly infers the next-best-view in a continuous space using a data-driven approach for the 3D object reconstruction task. We have validated the proposed approach making use of two groups of experiments. In the first group, several variants of the proposed architecture are analyzed. Predicted next-best-views were observed to be closely positioned to the ground truth. In the second group of experiments, the proposed approach is requested to reconstruct several unseen objects, namely, objects not considered by the 3D CNN during training nor validation. Coverage percentages of up to 90 \% were observed. With respect to current state-of-the-art methods, the proposed approach improves the performance of previous next-best-view classification approaches and it is quite fast in running time (3 frames per second), given that it does not compute the expensive ray tracing required by previous information metrics.
	\end{abstract}

	\section{Introduction}
	
	Automated three-dimensional (3D) object reconstruction or inspection is the process of building a 3D representation of a physical object by means of sensing its surface \cite{Scott_03}; its recent applications include inspection of airplanes \cite{jovanvcevic2015automated} or the reconstruction of heritage sites \cite{themistocleous2015methodology}. Due to the limited field of view of current sensors and incomplete models generated by single view reconstructions (in particular for non common objects like antique objects or art sculptures), the 3D models need to be completed by placing a visual or range sensor at several locations while the information is integrated into a partial model. 
	
	While state-of-the-art techniques for surface sensing and model integration are already mature enough for the task, for example, in Visual Simultaneous Localization and Mapping techniques \cite{murTRO2015} \cite{martinez2013enhancing}, the search of the optimal sensing locations remains as an open problem with growing interest by the robotics and computer vision community \cite{chen2011active}.
	
	Early work has defined the aforementioned problem as the computation of the next-best-views (NBV) \cite{Connolly_icra85}, where each NBV is the sensor view (position and orientation) that maximizes the reconstructed surface while the positioning and registration constraints are satisfied \cite{Scott_03}. Current methods can be classified into search-based or surface-based methods. In search-based methods, a set of candidate views is generated and then evaluated by a utility function \cite{delmerico2018comparison}\cite{vasquez2017view} \cite{doumanoglou2016recovering}. On the other hand, in surface-based methods, the reconstructed surface is analyzed to determine the NBV \cite{monica2018contour}\cite{chen2005vision}. Such methods require large computation times (search-based) or they are limited by object's auto occlusions (surface-based). \rtxt{Recent paradigms, in \cite{mendoza2018paper} and \cite{zeng2020pc},} have addressed the problem of NBV planning as a supervised learning problem where the objective is to find a function that predicts the NBV using previous knowledge. In the case of \cite{mendoza2018paper}, we have proposed a method for generating datasets and a 3D convolutional neural network (3D-CNN), called NBV-Net, for predicting the \btxt{position of the} NBV. \rtxt{However, such previous methods} are restricted to a classification, namely, the output of the 3D-CNN is limited to a small set of possible sensor views. In a real reconstruction case, such limitation could lead to an incomplete model. Therefore, it is necessary to obtain the NBV in a continuous domain.
	\btxt{ Unlike previous approaches, this paper presents further progress that is summarized as follows:
		\begin{enumerate}
			\item  The works \rtxt{in \cite{mendoza2018paper} and \cite{zeng2020pc}} address a classification problem, while the current paper solves a regression problem. The main implication is that in this work, one is not limited to discrete predefined sensing locations; instead, the NBV is determined in the continuum.
			\item The present paper provides an analysis in the terms of the number of layers in the network architecture. Such an analysis is not present in \cite{mendoza2018paper}.
			\item This work also presents an analysis in terms of the presence of dropout.
			\item A qualitative and quantitative analysis of predicted NBVs compared with their ground truth is provided. 
			\item We tested the method with thirteen new objects included neither in the training dataset \cite{nbvregdataset} nor in \cite{mendoza2018paper}.
			%\item We also consider more objects in the evaluation of the proposed approach compared with \cite{mendoza2018paper}.
		\end{enumerate}
	}
	
	\begin{figure}[t]
		\centering
		\includegraphics[width=0.9\linewidth]{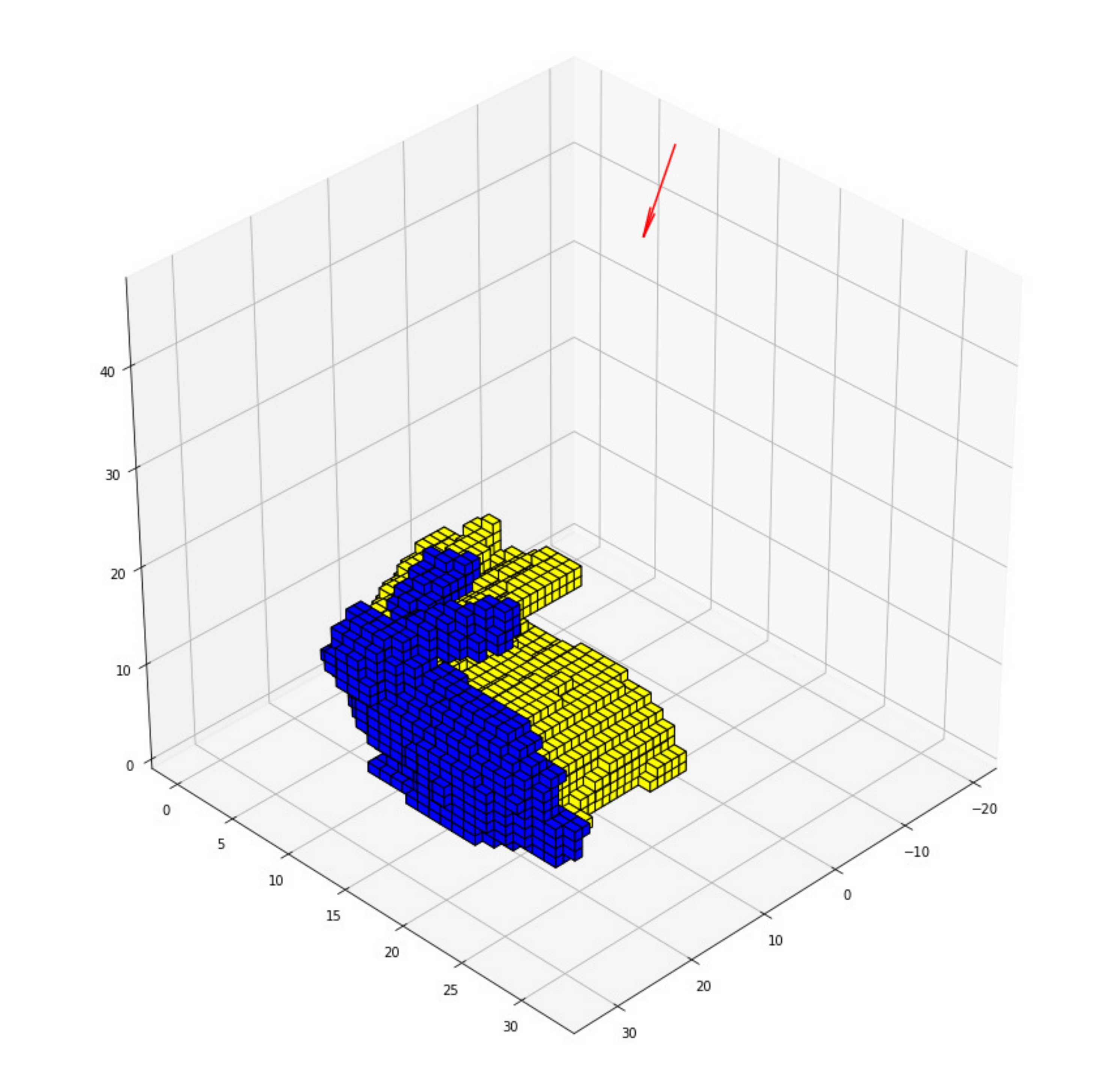}
		\caption{Example of a predicted next-best-view. The partial model is represented by a probabilistic grid where the blue voxels indicate scanned surface and yellow voxels indicate unknown space. The predicted next-best-view is drawn in red. We can see that from its position it will observe unknown volumes while maintains an overlap with already scanned surface. Figure best seen in color.}
		\label{fig:mainregression}
	\end{figure}
	
	As it was just mentioned, in this paper we propose a method for computing the NBV in a continuous domain. We have modeled the task as a regression problem where we want to learn a continuous function that receives as input the current state of the object reconstruction and predicts the NBV. To achieve the objective, we are using an extended dataset (with respect to \cite{mendoza2018paper}) and we are presenting a new CNN architecture. \btxt{The proposed CNN regresses the position of the NBV, while the orientation is computed geometrically.} We have validated the network performance using a test set and we have obtained a relatively small absolute error with respect to the ground truth NBV. In addition, the trained and validated network has been used for providing each sensor location during the reconstruction of completely unseen objects. The unseen objects are entities that were not seen by the network during training nor validation. As a result, the proposed approach has been capable of providing each sensor location in the continuous six-dimensional space (position and rotation). See Fig. \ref{fig:mainregression} as an example of a computed NBV. With respect to the current approaches, the proposed method is quite fast, one third of a second, because it avoids the ray tracing step, required by the majority of current state-of-the-art methods \cite{delmerico2018comparison}\cite{vasquez2017view}, and it only requires a forward pass of the network.

	\btxt{
		A possible application of this work is verification of the size and shape of a given object. That is, the automatic modelling can be used to determine how well the object matches the design specifications. Another application is to use the model as an input to perform an automatic manipulation task with a robot. Moreover, the fast performance achieved in the present approach is an important step toward online NBV computation \cite{song2017online}, in which new gathered information is used for fast NBV recalculation as the sensor moves. \rtxt{Decreasing the computation time of the NBV gives rise to a series of potential benefits, for instance, it allows one to execute complex tasks such as simultaneous object reconstruction and manipulation, fast building modeling and inspection which may be critical for human safety in emergency situations such as earthquakes and floods, reactive obstacle modelling and collision avoidance for reactive navigation, among others.} 
		}

	\subsection{Comparison with related methods and paper organization} 
	
	In order to show the benefits and drawbacks of the proposed method, a comparison, in terms of percentage of object reconstruction and processing time needed to compute the next-best-view, between the method proposed in this work and other two related approaches will be presented. One of the methods is based on classification 
	\cite{mendoza2018paper} and the other uses information gain to exhaustively evaluate a given number of views \cite{kriegel2012next}.

The rest of the paper is organized as follows. \btxt{Section \ref{sec:related_work} provides a brief overview of relevant recent work on the field.} Section \ref{sec:mat_and_met} provides the background of the proposed method, including the description of the used dataset. Section \ref{sec:regression} presents the proposed method for addressing the problem. Section \ref{sec:experiments} describes the experiments, including networks training as well as the reconstruction of unseen objects. Finally, section \ref{sec:conclusions} gives the conclusions and future research directions.

\section{Related work}
\label{sec:related_work}
	\btxt{
		As we have already mentioned, the present paper is about the computation of the next-best-view (NBV) for 3D object  reconstruction. There is a lot of work available in this area. In  \cite{Scott_03,chen2011active,zeng2020view} interesting surveys about object reconstruction 
		are presented.} 
	\btxt{ 
		Some relevant works among many contributions are the following.
		The work  in \cite{Torabi_ijrr12} presents a method for
		planning a next-best-view for object reconstruction in the workspace.
		The approach uses inverse kinematics computation to get a configuration matching 
		the desired sensor location.  In \cite{Kriegel13}, the authors have as a main objective to obtain a high
		quality surface model allowing applications such as
		grasping and manipulation. That work integrates 3D modeling, autonomous 
		view planning and motion planning in a coherent manner.}
	
	\btxt{
		The authors in \cite{Kriegel13} use  Probabilistic Road Maps (PRMs) \cite{Kavraki_96}  and
		Rapidly-Exploring Random Trees (RRTs) \cite{LaValleRRT}  to find collision
		free paths. In~\cite{Khalfaoui13}, the authors propose a method  to determine the
		next-best-view for an efficient reconstruction of highly
		accurate 3D models. The approach is based on the classification of the
		acquired surfaces, it also  combines that classification  with a
		best view selection algorithm based on mean shift.
		In \cite{Potthast_14}, the authors present an information gain-based variant of NBV problem 
		for a cluttered environment. They propose a belief model that allows one to obtain an accurate 
		prediction of the potential information gain of new viewing locations.
		In \cite{delmerico2018comparison} the authors investigate which formulation of 
		information gain is best for a volumetric 3D reconstruction with
		a robot, which is equipped with a dense depth sensor. The authors also provide a
		comparative study about the performance of information gain metrics
		for active 3D object reconstruction.}
	
	\btxt{ 
		In \cite{vasquez2017view}, the authors propose a method for next-best-view/state planning
		for 3D object reconstruction. The proposed method generates 
		a set of candidates in the state space, later only
		a subset of these views is kept by filtering the original set.
		A utility function that integrates several aspects of the problem and an efficient strategy to evaluate the candidate views is proposed.
		The work in \cite{lauri2020multi} addresses NBV planning for multiple depth cameras and propose a utility function that scores sets of view-points and avoids overlap between multiple sensors. The authors proved that multi-sensor NBV planning with such utility function is an instance of submodular maximization under a matroid constraint, allowing them to propose a polynomial-time greedy algorithm.}
	
	\btxt{
		In \cite{song2017online},  a method is proposed for inspecting a
		partially known environment. First,  a target goal is computed
		and then it determines a path until a local area is inspected.
		The inspection is declared as  complete if the percentage of unknown volume with
		respect to the whole unknown volume is lower than a threshold.
		In \cite{Song-18}, the same authors mentioned that
		the completion of a volumetric map does not necessarily
		describe the completion of a 3D model, consequently, the model completeness is evaluated, 
		according to the quality of the reconstructed  surfaces. 
		Also concerning exploration and inspection of 3D-environments, the authors of \cite{hardouin2020surface} utilize a map representation based on a Truncated Signed Distance Function (TSDF). The TSDF data is used to identify missing parts of the model and generate a list of candidate sensor configurations, the visit of which is scheduled using an NBV planning method.} 
	
	\btxt{In \cite{Srini-18} a motion planning strategy for exploration of ground-level structures is proposed. 
		The method has two main steps, in the first step, it follows the contour of an
		unknown target, then the robot moves to the missing portions of the reconstructed model. The work in \cite{moritani2020plausible} consider the 3D model reconstruction problem in the context of infrastructure maintenance. Their approach first reconstructs an approximate 3D model using only sparse point clouds generated from a Structure-from-Motion method; the resulting rough model can be used to predict the quality of the final dense model and for an optimization-based view planning based on degraded regions.}
	
	\textcolor{black}{
		On the other hand, there are some related methods based on machine learning. \rtxt{The majority address view planning for object recognition. For example, Wu et al. \cite{wu20153d} proposed a deep belief network to perform surface prediction for model completion in order to achieve a faster object recognition. Such predicted surface is used to guide the next sensing location.} Johns et al. \cite{johns2016pairwise} apply deep learning for computing the sequence of views that increase the mutual probability of recognizing an object. Their method uses pairs of views such that the problem become tractable. Their method uses two neural networks, one for predicting the object class and one for predicting the next-best-view from a predefined view sphere. Even-though they show promising results, the views are still limited to a given set and their proposed method can not be directly applied to object reconstruction. In consequence, for dealing with object reconstruction, Hepp et al. \cite{hepp2018learn} proposes to learn a function that measures the utility of a candidate view. The supervised learning process uses known volumetric maps to determine the ground truth utility. Bai. \cite{bai2017toward} proposes an exploration method based on a deep neural network for selecting the robot 2d position from a discrete set of positions. The exploration method determines the best position in two steps: first, the network outputs some candidate positions, then, those candidates are
		evaluated by Mutual Information \cite{julian2014mutual}. The strategy proposed by Wang et al. \cite{wang2018autonomous} combines a hand-crafted utility with a learned utility. The learned utility is structured as a classification approach with a CNN based on AlexNet. Inside it, the input is a range image and the outputs is a vector with the scores of a fixed set of views.  \rtxt{In~\cite{wu2019plant}, a NBV method is proposed that leverages deep learning for phenotyping plants. The approach uses a network based on the Point Completion Network (PCN)~\cite{yuan2018pcn}, but with the capability to predict the confidences of completed points. The network learns the structural prior of plants, receives cloud points that partially model the plant in question, and outputs a completed point cloud of the plant. Subsequently, the resulting point cloud is used to build a predicted octomap model, which in turn is used to guide NBV planning; the next-best viewpoint is defined as the one that can provide the maximum amount of information for the plant phenotyping.} \rtxt{Zeng et al. \cite{zeng2020pc} proposes a deep neural network for evaluating a set of candidate views efficiently. Their method, contrary to \cite{mendoza2018paper}, require as input the reconstructed point cloud, avoiding the need of an intermediate representation such as the probabilistic grid. Zeng et al. architecture provides efficient evaluation. However, their method is still limited to a predefined set of views.}
		Therefore, to the best of our knowledge, there is no method for estimating directly the position of the next-best-view in the 3D space for 3D object reconstruction.
	}
	
	\section{Background}
	\label{sec:mat_and_met}
	
	In this section, we provide the background of the proposed method for learning the next-best-view.
	
	\subsection{3D reconstruction}
	
	The automated 3D reconstruction is an iterative process of four steps: positioning, sensing, registration-and-update and next-best-view planning. The positioning places the sensor at the desired pose, then the surface is measured by the sensor, next, the registration-and-update transforms the observed surface to a common reference frame \cite{icp} and integrates the information into a single partial model \cite{hornung13auro}, then the next-best-view planning determines the next sensor pose. The process is repeated until an stop condition.
	
	Some assumptions that are made in this paper are the following: the object of interest is encapsulated in a cube, the center of the cube is assumed as the center of the object, the sensor is capable of obtaining a point cloud from the object's surface.
	
	\subsection{Partial model}
	
	The partial model, $\mathcal{M}$, stores the accumulated information about the object of interest. In this paper, it is implemented with a probabilistic grid where the space is evenly divided into small cubes. Each cube is called voxel and has associated a probability of being occupied. The occupancy probability is updated with a Bayes filter using the perceptions from the sensor \cite{Thrun_book}. Considering $m$, $n$ and $o$ as the dimensions of the grid, the partial model will also be written as $\mathbb{R}^{m \times n \times o}$. A reconstruction state is an instance of the partial model; two partial models can contain the same object but they can differ in the reconstruction state. In this paper, we are using the Octomap library \cite{hornung13auro} to implement the probabilistic grid.
	
	\subsection{Next-best-view}
	
	The next-best-view is defined as a six-tuple:
	
	\begin{equation}
	v = (x,y,z,\alpha, \beta, \gamma)
	\end{equation} where $x$, $y$ and $z$ define a position in the 3D euclidean space and $\alpha$, $\beta$ and $\gamma$ define the yaw, pitch and roll orientations according to the Tait-Bryan angles. Based on the orientations, a rotation matrix, $R(\alpha, \beta, \gamma)$, that transforms the sensor pose can be directly obtained
	from the rotation angles. 
	
	\subsection{Dataset}
	
	The dataset that we are using was proposed in a previous work from our research group \cite{mendoza2018paper}. This dataset contains tuples of regressors and responsors. The regressor is a partial model, $\mathcal{M}_i \in \mathbb{R}^{32 \times 32 \times 32}$, and the responsor is its corresponding next-best-view, $v_i = \left( x,y,z,\alpha, \beta, \gamma \right) $, where $i$ is an index over the dataset. The dataset was build by performing several reconstructions for different objects. The reconstruction scene places the object at the origin of the global reference frame and a large set of possible sensor locations are generated forming a sphere of radius 0.4 m. 12 object shapes were included in the dataset. During the reconstructions, for each partial model a next-best-view was computed by performing an exhaustive search over the set of possible views. The view that maximized the increment of reconstructed surface and satisfy an overlap was taken as the ground truth NBV. In our previous work \cite{mendoza2018paper}, we restrict possible predictions to a finite set of views (14 classes) around the object. Unlike it, in this paper, we are not restricting the predictions, therefore the possible predictions are in the 6D space. The dataset that we are using in this paper is available at \cite{nbvregdataset} and the reduced dataset used in our previous work is available at \cite{nbvclassdataset}.

	\begin{figure}[tb]
		\centering
		\includegraphics[width=\linewidth]{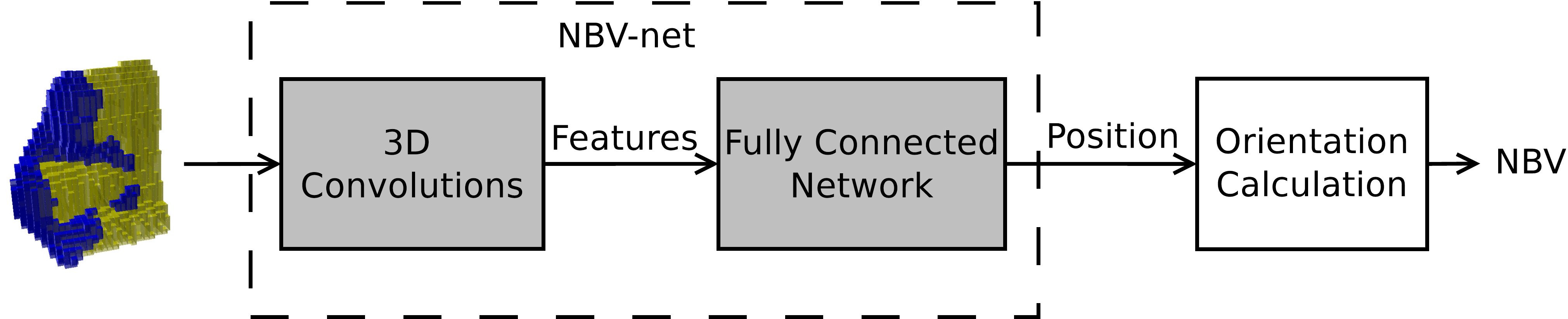}
		\caption{Overall regression approach for next-best-view planning.}
		\label{fig:approach} 
	\end{figure}
	
	\section{Next-best-view regression}
	\label{sec:regression}
	
	The next-best-view regression is the task of finding a function \begin{equation}
	f(\mathcal{M}):\mathbb{R}^{m \times n \times o} \rightarrow \mathbb{R}^{3} \times SO(3)
	\label{eq:f_learning}
	\end{equation} so that, the information perceived in $\hat{v} = f(\mathcal{M})$ increases the surface of the object contained in the partial model $\mathcal{M}$ satisfying the registration and positioning constraints \cite{mendoza2018paper}. Notice that the input  of $f$ is the probabilistic grid where $m \times n \times o$ is the number of voxels and the output is composed by a position in $\mathbb{R}^{3}$ and a orientation represented by a rotation matrix in the special orthogonal group $SO(3)$.

	\subsection{Regression approach}

	In the proposed scheme, we use a 3D convolutional neural network, denoted by $\Phi$, to regress the position of the NBV. Then the orientation is computed by aligning the sensor to the center of the object. This approach has the advantage of being easier to train, given that only three continuous variables, corresponding to the position, are predicted by $\Phi$, while the orientation is computed using geometric reasoning. Fig. \ref{fig:approach} depicts the whole approach to compute the NBV. Formally, the network output is:
	
	\begin{equation}
	\hat{p} = \Phi(\mathcal{M})
	\label{eq:predicted_position}
	\end{equation} where $\hat{p} = ( x_{\hat{p}}, y_{\hat{p}}, z_{\hat{p}} )$ is a position in $\mathbb{R}^3$ where the sensor will be placed. \textcolor{black}{Given that in running time the objects can have different sizes w.r.t. the training samples, the predicted position ($\hat{p}$) must be scaled in order to fit the object size. This is achieved by maintaining the same grid resolution ($32 \times 32 \times 32$) but changing the voxel size according to the object size and scaling the predicted position by a factor $k$}:
	\begin{equation}
	s = k \cdot \hat{p}
	\end{equation} \textcolor{black}{The $k$ factor can be calculated so that the sensor’s field-of-view encloses the object of interest, that is, given the smallest sensor’s opening angle and the object’s major span, the $k$ factor is the distance such that the object lies within such an opening angle.}
	
	Once the position is predicted by the network, the orientation is computed first as a unit vector, \\ $\hat{r} = \left(x_r,y_r,z_r \right)$, indicating the orientation of the sensor's director ray, namely:
	
	\begin{equation}
	\hat{r} = \frac{c-\hat{p}}{||c-\hat{p}||} 
	\end{equation} \textcolor{black}{where $c$ is the center of the object}. Then, $\hat{r}$ is converted to Euler rotation angles:
	
	\begin{equation}
	\alpha = \arctan \left( \frac{y_r}{x_r} \right) 
	\end{equation}
	
	\begin{equation}
	\beta = \arcsin\left( z_r \right) 
	\end{equation}
	
	It is worth to say that under this approach, $\hat{r}$ provides the yaw and pitch parameters but the roll parameter (rotation over the camera axis) is omitted. In consequence, $\gamma = 0$. Even though one degree of freedom is lost, the roll angle is usually omitted in next-best-view planning because it has the lowest impact on the built model. Finally, the predicted NBV is given by:
	
	\begin{equation}
	\hat{v} = \left( s_x, s_y, s_z, \alpha, \beta, \gamma \right)
	\end{equation}

	\subsection{NBV-net}
	\label{sec:nbv_net}
	
	\begin{figure}[tb]
		\centering
		\includegraphics[height=\linewidth, angle =90]{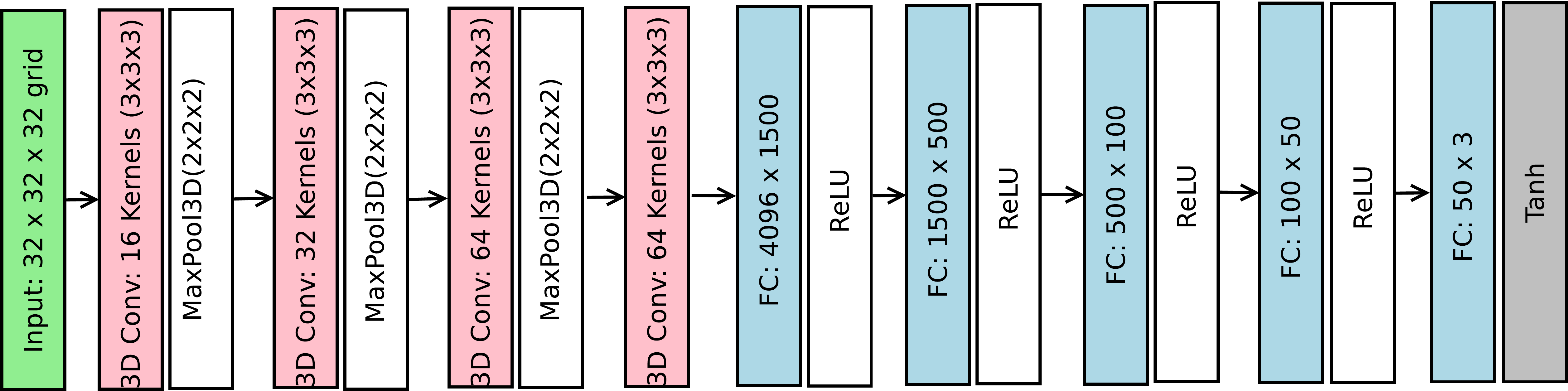}
		\caption{ NBV-net 4-5 architecture. The number 4-5 stands for 4 feature extraction layers and 5 fully connected layers.}
		\label{fig:nbvnetplus}
	\end{figure}

	In a previous work \cite{mendoza2018paper}, our research group presented a 3D CNN for view classification. However, it has shown poor performance for the regression task, as we will discuss later in the experiments.  Therefore, we propose to extend our previous network to the regression task.

	The proposed architecture, NBV-net 4-5, receives as input a probabilistic grid of dimension $32 \times 32 \times 32$ and predicts the position of the NBV. It has four 3D convolutional layers and five connected layers (giving the name 4-5). The 3D convolutional layers extract features from the grid to a 4D vector, then the features are flattened and passed through the set of fully connected layers. See Fig. \ref{fig:nbvnetplus}.

	To simplify the detailed network description, we will use the following notation: $C(f,k,s)$ defines a 3D convolutional layer with $f$ filters of size $k \times k \times k$ and stride $s \times s \times s$, $P(s)$ a max pooling layer of stride $s \times s \times s$ and $FC(n)$ defines a fully connected layer with $n$ nodes.

	Then, NBV-net 4-5 is configured as follows: C(16,3,2), P(2), C(32,3,2), P(2), C(64,3,2), P(2), C(64,3,2), FC(1500), FC(500), FC(100), FC(50), FC(3). Note that, there is no pooling after the fourth convolutional layer and the 4096 features are pass through three fully connected layers. The activations are performed by Rectified Linear Units (ReLU) except for the last one which applies a hyperbolic tangent activation function (Tanh).

	\begin{figure*}[t]
		\centering
		\begin{subfigure}[b]{0.24\textwidth}
			\includegraphics[width=\textwidth]{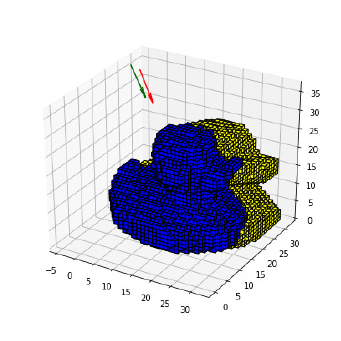}
			\caption{Toy duck}
			\label{fig:duck}
		\end{subfigure}
		\begin{subfigure}[b]{0.24\textwidth}
			\includegraphics[width=\textwidth]{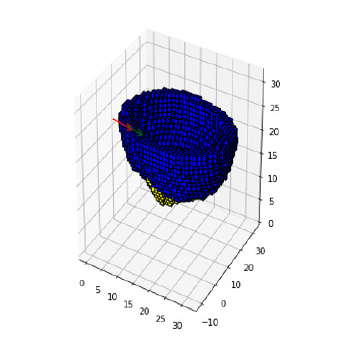}
			\caption{Bowl}
			\label{fig:bowl}
		\end{subfigure}
		\begin{subfigure}[b]{0.24\textwidth}
			\includegraphics[width=\textwidth]{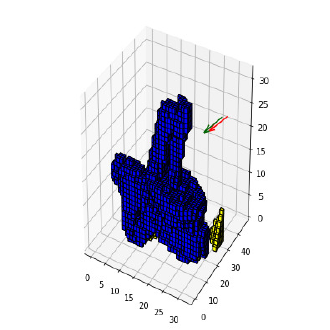}
			\caption{Work bench}
			\label{fig:bech}
		\end{subfigure}
		\begin{subfigure}[b]{0.24\textwidth}
			\includegraphics[width=\textwidth]{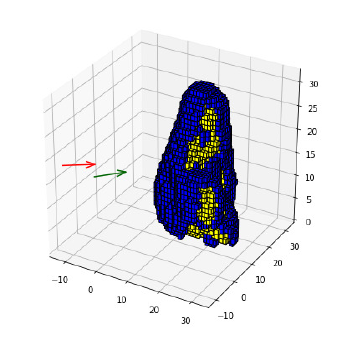}
			\caption{Monkey}
			\label{fig:monkey}
		\end{subfigure}
		\begin{subfigure}[b]{0.24\textwidth}
			\includegraphics[width=\textwidth]{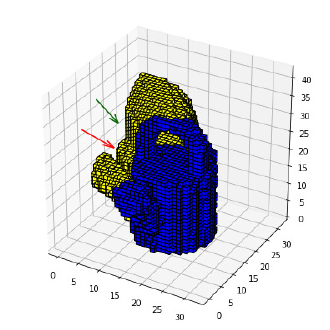}
			\caption{Plant sprinkler}
			\label{fig:sprinkler}
		\end{subfigure}
		\begin{subfigure}[b]{0.24\textwidth}
			\includegraphics[width=\textwidth]{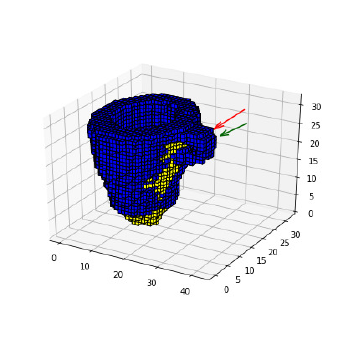}
			\caption{Cup}
			\label{fig:cup}
		\end{subfigure}
		%	~ %add desired spacing between images, e. g. ~, \quad, \qquad, \hfill etc. 
		\begin{subfigure}[b]{0.24\textwidth}
			\includegraphics[width=\textwidth]{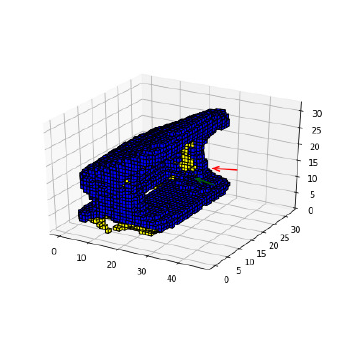}
			\caption{Paper punch}
			\label{fig:punch}
		\end{subfigure}
		\begin{subfigure}[b]{0.24\textwidth}
			\includegraphics[width=\textwidth]{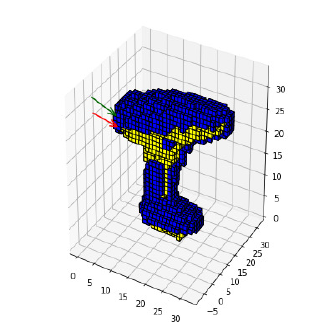}
			\caption{Drill}
			\label{fig:drill}
		\end{subfigure}
		\begin{subfigure}[b]{0.24\textwidth}
			\includegraphics[width=\textwidth]{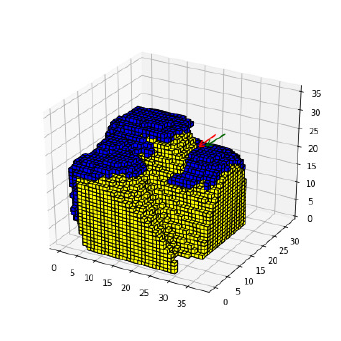}
			\caption{Camera}
			\label{fig:camera}
		\end{subfigure}
		\begin{subfigure}[b]{0.24\textwidth}
			\includegraphics[width=\textwidth]{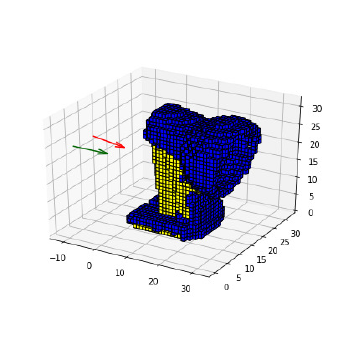}
			\caption{Desk Lamp}
			\label{fig:lamp}
		\end{subfigure}
		\begin{subfigure}[b]{0.24\textwidth}
			\includegraphics[width=\textwidth]{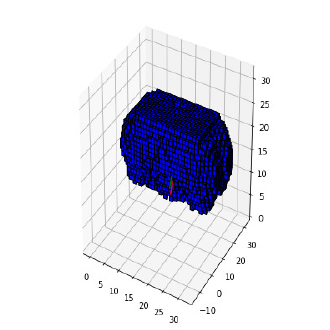}
			\caption{Egg carton}
			\label{fig:carton}
		\end{subfigure}
		\begin{subfigure}[b]{0.24\textwidth}
			\includegraphics[width=\textwidth]{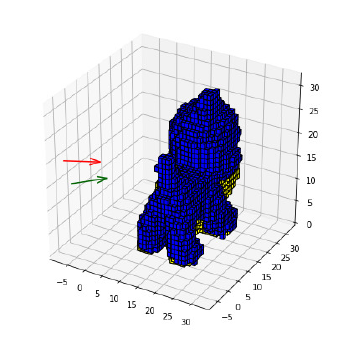}
			\caption{Toy Cat}
			\label{fig:cat}
		\end{subfigure}
		
		\caption{Comparison of the predicted next-best-view versus the ground truth for several objects in the dataset. Blue voxels indicate measured surface. Yellow voxels indicate unknown space. The predicted next-best-view is drawn in red. The ground truth next-best-view is drawn in green. Figure best seen in color.}
		\label{fig:predictions}
	\end{figure*}

	\section{Experiments}
	\label{sec:experiments}
	
	The main reason for selecting a particular CNN architecture or a set of hyper-parameters is to obtain a good generalization on the task. In this context, a good generalization will be to provide a NBV that increases the object surface despite the variability of i) the object shape and ii) the current reconstruction state. Fig. \ref{fig:predictions} can be helpful to observe the different object shapes and reconstruction states. In that sense, the dataset \cite{nbvregdataset} provides more than ten thousands of examples of reconstruction states because the objects were reconstructed several times from different initial positions. However, the shapes are limited to 12. For this reason, the experiments will focus on showing the network's architecture and parameters that provide a better generalization. 
	
	We present two groups of experiments. The first group analyzes the training and validation of several variants of the proposed architecture. The second group of experiments test the most promising nework configurations in the reconstruction of several unseen objects, these objects were not seen by the CNN during training nor validation. At the end of this section, we present a discussion about the network advantages as well as the current challenges. For all the experiments, the architectures were implemented in PyTorch. The experiments were carried out using an Intel i7 machine with NVIDIA Geforce 1080 GPU.

	\subsection{Network training and architecture variants}

	In this group of experiments, we analyze the training and validation performance of the proposed architecture. First, we compare and analyze several variants of the architecture including those reported previously. Then, we analyze the use of the regularization method dropout during training.

	\subsubsection{Architecture variants}

	Several architectures for 3D regression problems have been proposed, for example to determine the pose of a known object. However, in NBV planning for 3D object reconstruction, the number of current methods is very limited. Therefore, we compare the proposed network versus three variants, where one of them is a modified version of \rtxt{an} architecture reported for classification.

	\begin{itemize}
		\item NBV-net 3-3. The shortest network; it includes three convolutional layers and 3 fully connected layers. In detail, C(10,3,2), P(2), C(12,3,2), P(2), C(8,3,2), P(2), FC(1024), FC(500), FC(3).
		\item NBV-net 3-5. Network proposed in \cite{mendoza2018paper} for NBV classification. The last layer is replaced by three nodes with a \textit{Tanh} function as the activation function. In detail, C(10,3,2), P(2), C(12,3,2), P(2), C(8,3,2), P(2), FC(1500), FC(500), FC(100), FC(50), FC(3).
		\item NBV-net 4-3. This variant increses the number of feature extraction layers to 4 but decreases the fully connected layers to 3. In detail, C(16,3,2), P(2), C(32,3,2), P(2), C(64,3,2), P(2), C(64,3,2), FC(1024), FC(500), FC(3).
		\item NBV-net 4-5. Both, feature extraction and fully connected layers are increased. Description presented in section \ref{sec:nbv_net}.
	\end{itemize}

	\begin{figure}[tb]
		\centering
		\includegraphics[width=\linewidth]{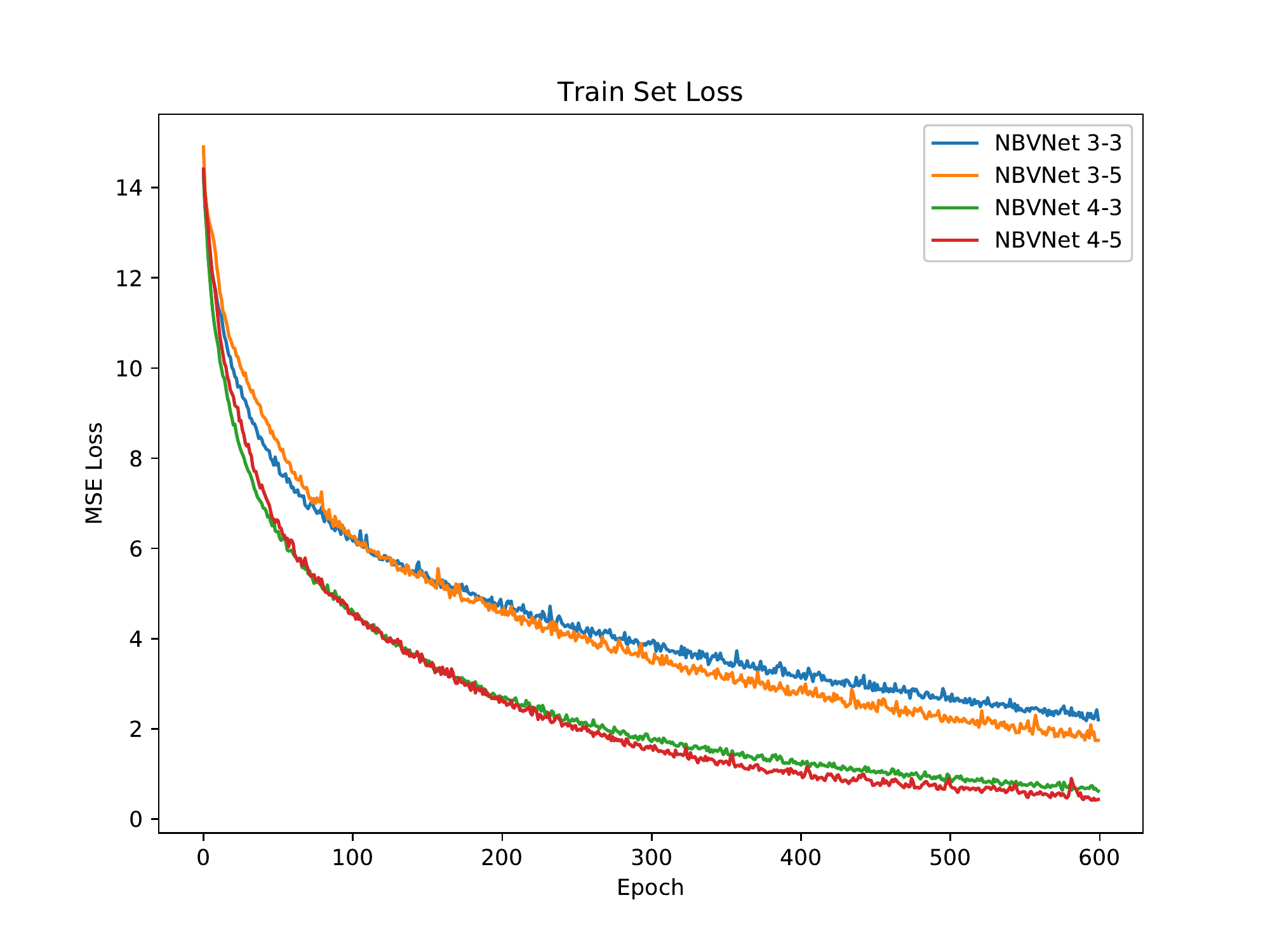}
		\caption{Training loss.}
		\label{fig:traininglossdrop}
	\end{figure}

	\begin{figure}[tb]
		\centering
		\includegraphics[width=\linewidth]{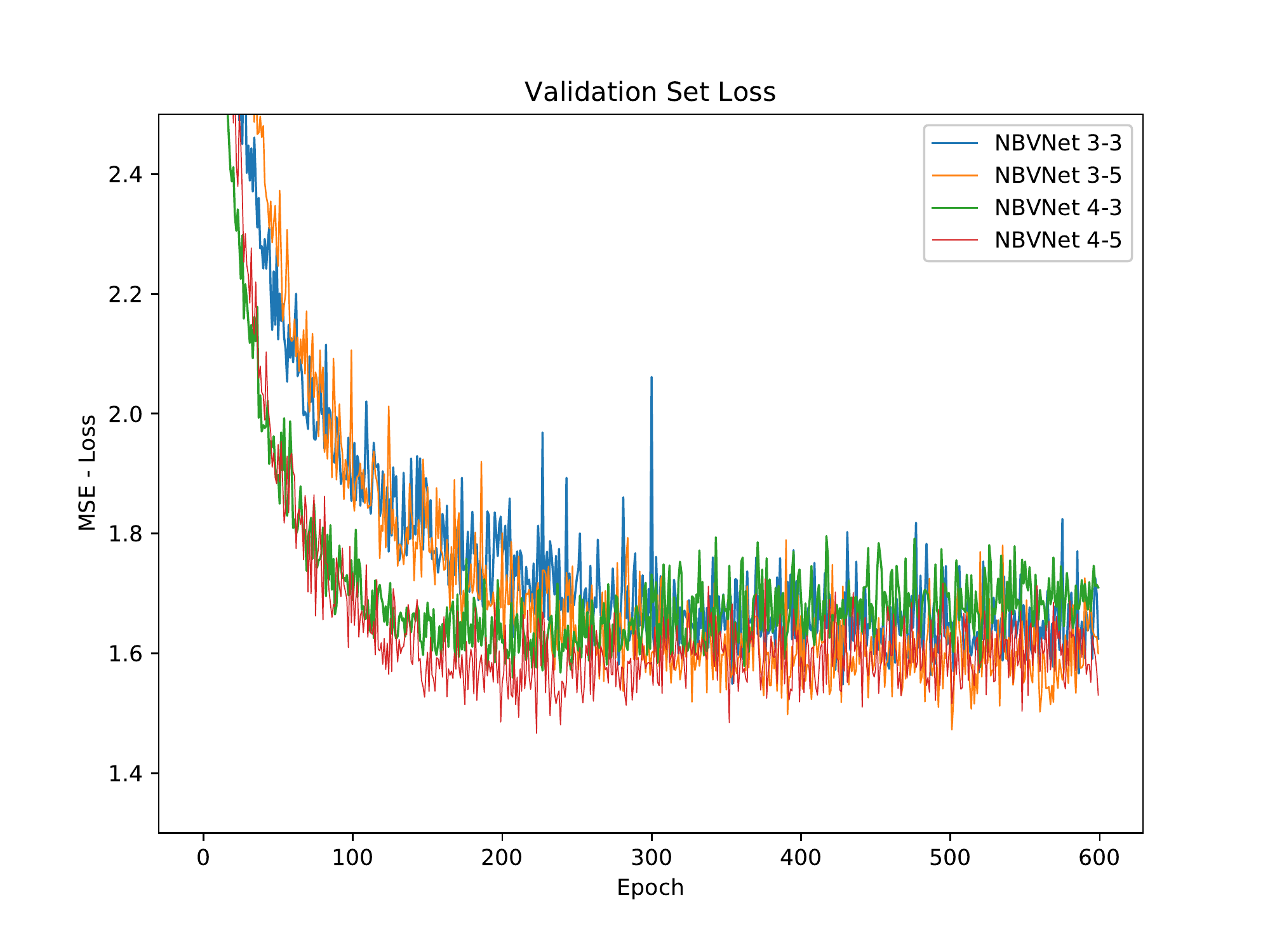}
		\caption{Validation loss.}
		\label{fig:validationlossnbv7}
	\end{figure}

	\begin{figure}[tb]
		\centering
		\includegraphics[width=\linewidth]{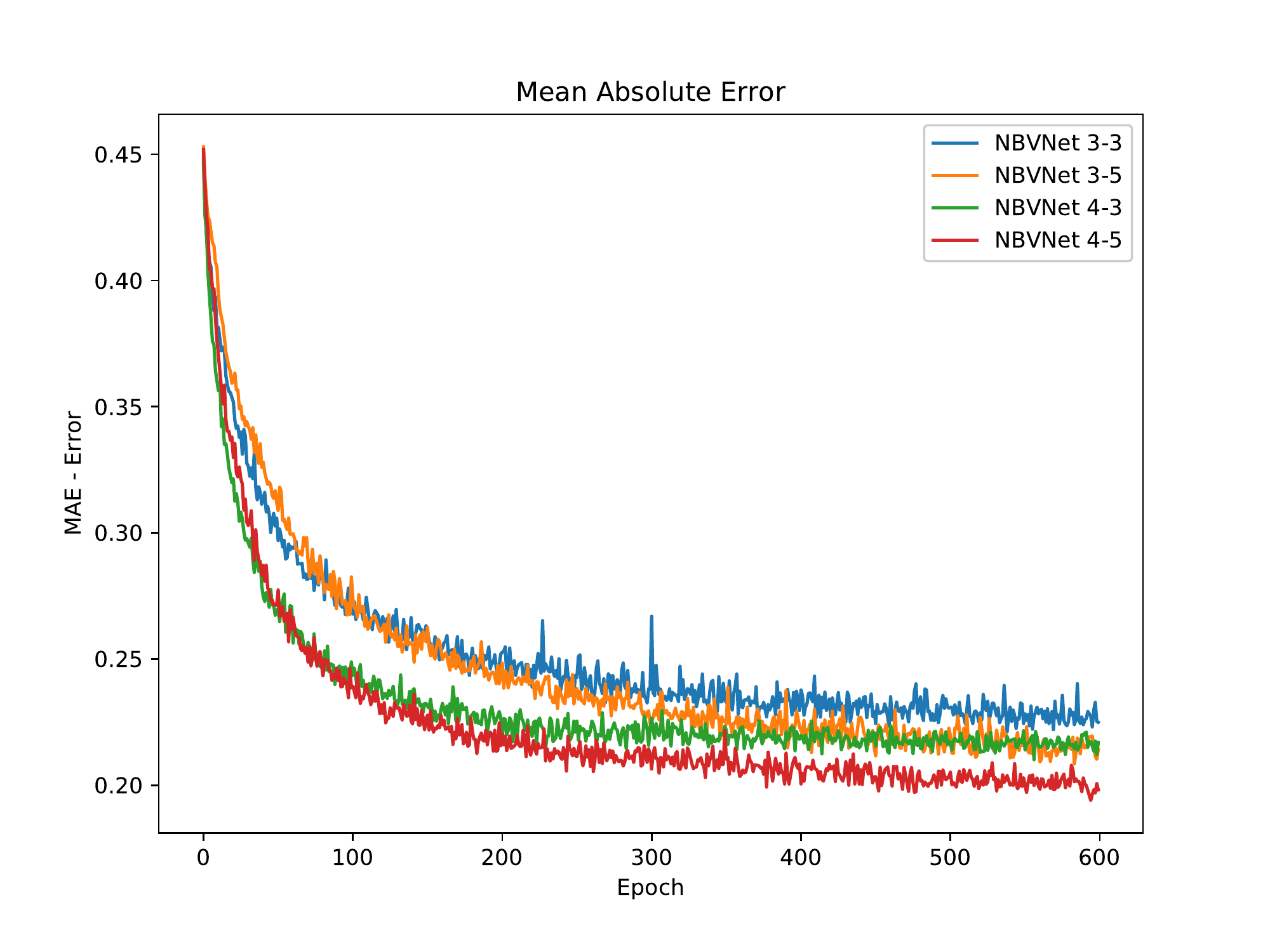}
		\caption{Mean absolute error (MAE) over validation set. The graph shows how far in the euclidean space are the predictions from the ground truth next-best-view.}
		\label{fig:maenbv7}
	\end{figure}
	
	\begin{figure}
		\centering
		\includegraphics[width=\linewidth]{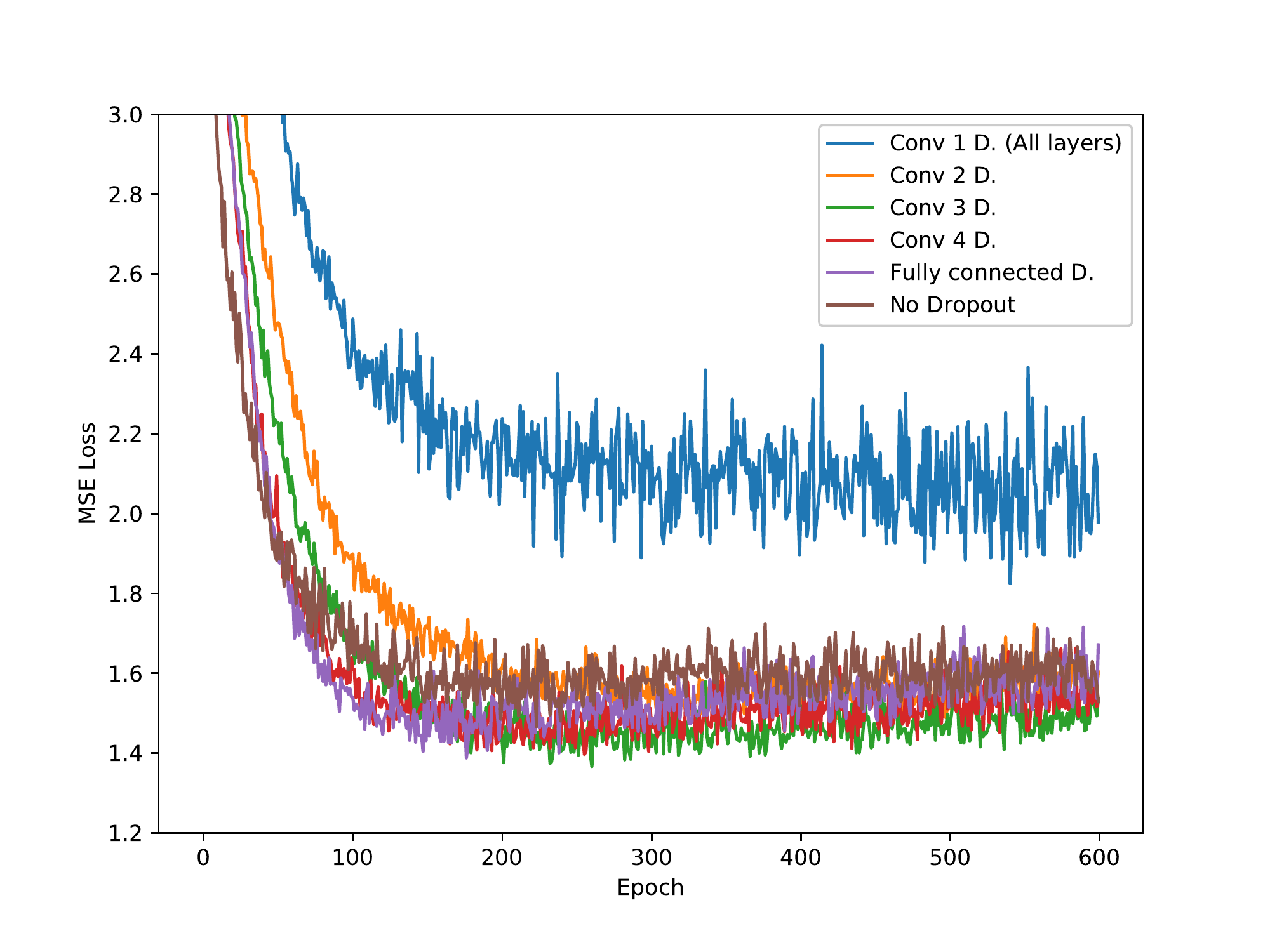}
		\caption{Mean square error (MSE) over validation set for different dropout configurations. The regularization method dropout is inserted at several network layers. Nomenclature ``Conv N D" indicates that dropout is included from convolutional layer N until the last network layer. Network tested: NBV-net 4-5.}
		\label{fig:maeregularization}
	\end{figure}
	
	\subsubsection{Training}
	
	According to the proposed model, we require to predict the NBV position, equation (\ref{eq:predicted_position}). Thus, the ground truth positions are normalized to unit vectors. Then, the networks are trained to reduce the Mean Squared Error (MSE) loss between prediction $\hat{p}$ and ground truth position $p$:
	
	\begin{equation}
	loss\left(p, \hat{p} \right) = \frac{1}{N}\sum_{i=1}^{N} \left( \hat{p}_i - p_i \right)^2 
	\label{eq:loss}
	\end{equation} where $N=3$ is the number of elements in the position vector. To minimize the error, we use the Adam optimizer \cite{kingma2014adam} which auto-adjusts the learning rate.
	
	The dataset was divided randomly into 80\% for training and 20\% for validation. In this way, both subsets contain examples from the 12 objects but with different reconstruction states, this imply a different distribution of the occupied, unknown and free voxels. The training was performed during 600 epochs with learning rate 0.0001 and batch size 250. Each network training required an average time of five hours.

	\subsubsection{Precision analysis}

	The training loss, calculated with eq. (\ref{eq:loss}), for the training set is shown in Fig. \ref{fig:traininglossdrop}. As we can see, all networks reduce the loss, but the best fitting ones, according to MSE, are the networks that include four convolutional layers instead of only three. On the other hand, with respect to the validation set, all networks reach a similar loss, see Fig. \ref{fig:validationlossnbv7}; however, we can observe that the networks with four convolutional layers start to overfit after 200 epochs, namely, the loss increases for such networks. For that reason, we stop growing the depth of the proposed network. Some examples of the predictions made by NBV-net 4-5 are shown in Fig. \ref{fig:predictions}.
	
	Previous analysis focus on training performance, however, in order to provide additional information about how well in the Euclidean space the predictions are, Fig. \ref{fig:maenbv7} shows the  mean absolute error (MAE) as the training epochs advance. Note that MAE was not used for training. 
	
	\begin{equation}
	MAE( p, \hat{p} ) = \frac{1}{N}\sum_{i}^N |\hat{p}_i - p_i |
	\end{equation}

	We can observe from Fig. \ref{fig:maenbv7}, that NBV-net 4-5 was the one with the best result, in terms of the smallest Euclidean distance between the predictions and the ground truth next-best-view. NBV-net 4-5  is the architecture that has the largest number of convolutional and fully connected layers.  We can also observe that NBV-net 4-3 has also a good performance, particularly in the first epochs, where it has better results than NBV-net 4-5.
	
	\subsubsection{Regularization}
	
	Regularization is used for avoiding overfitting over the training set. In this context, we want to prevent learning the object’s shapes in the dataset. Therefore, in this experiment we analyze how the regularization dropout method affects network performance. The experiment inserts dropout with probability 0.5 layer by layer, starting from the fully connected layer until the first convolutional. Fig. \ref{fig:maeregularization} shows the MSE over the validation set using the NBV-net 4-5. As we can see in the graph, after 600 epochs, inserting dropout from the third convolutional layer until the last one (Conv 3 D) leads to the smallest MSE loss, meaning that even though some intermediate features are not present the output is adequate. However, if dropout is applied from the first convolutional layer (Conv 1 D.), then the MSE increases dramatically. Our hypothesis for this phenomenon is that low level feature extraction is very important, therefore by missing any of such features the NBV regression is affected.
	
		\begin{figure*}[tb]
		\centering
		\begin{subfigure}{0.19\linewidth}
			\includegraphics[width=\textwidth]{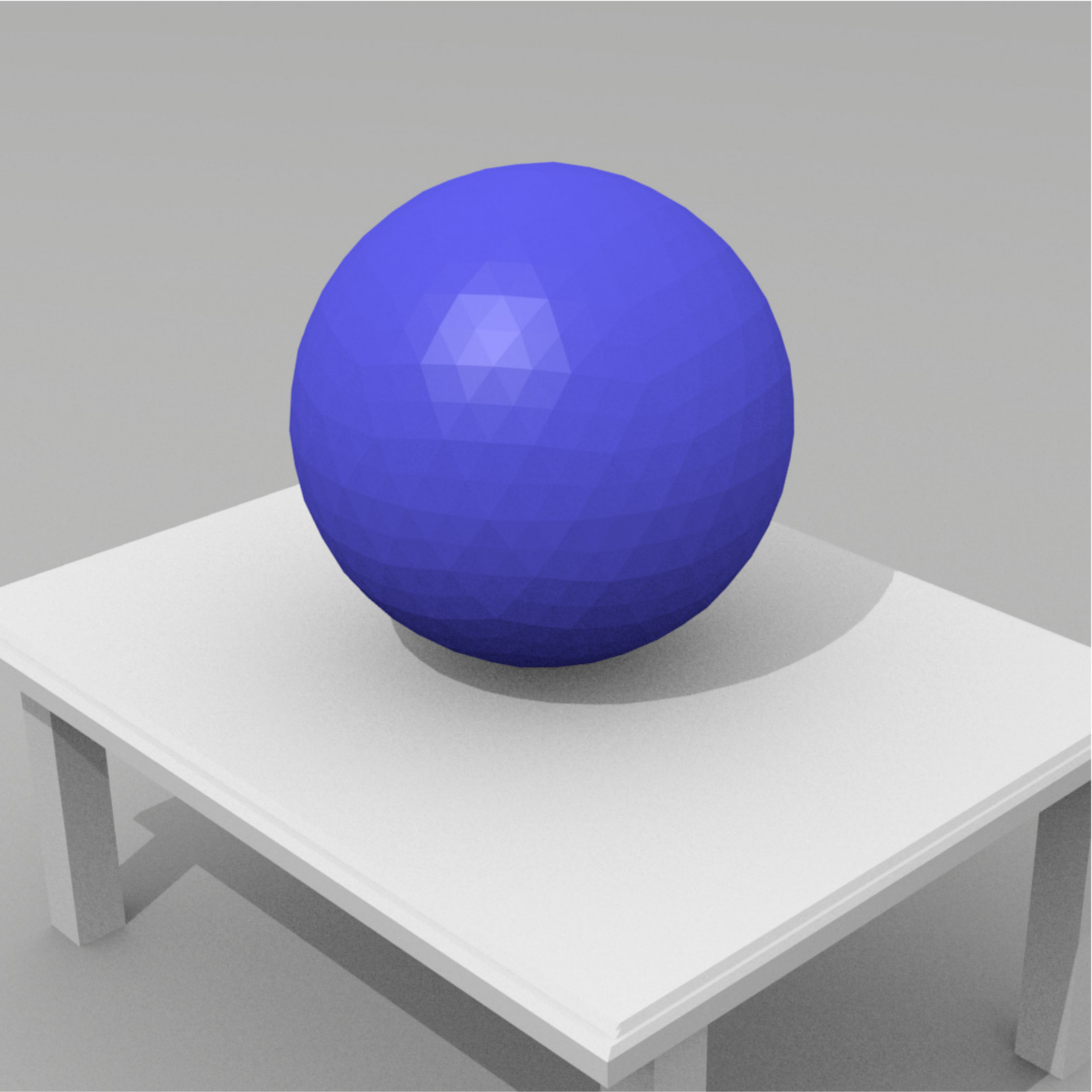}
			\caption{Sphere}
			\label{fig:sphere}
		\end{subfigure}
		\begin{subfigure}{0.19\linewidth}
			\includegraphics[width=\textwidth]{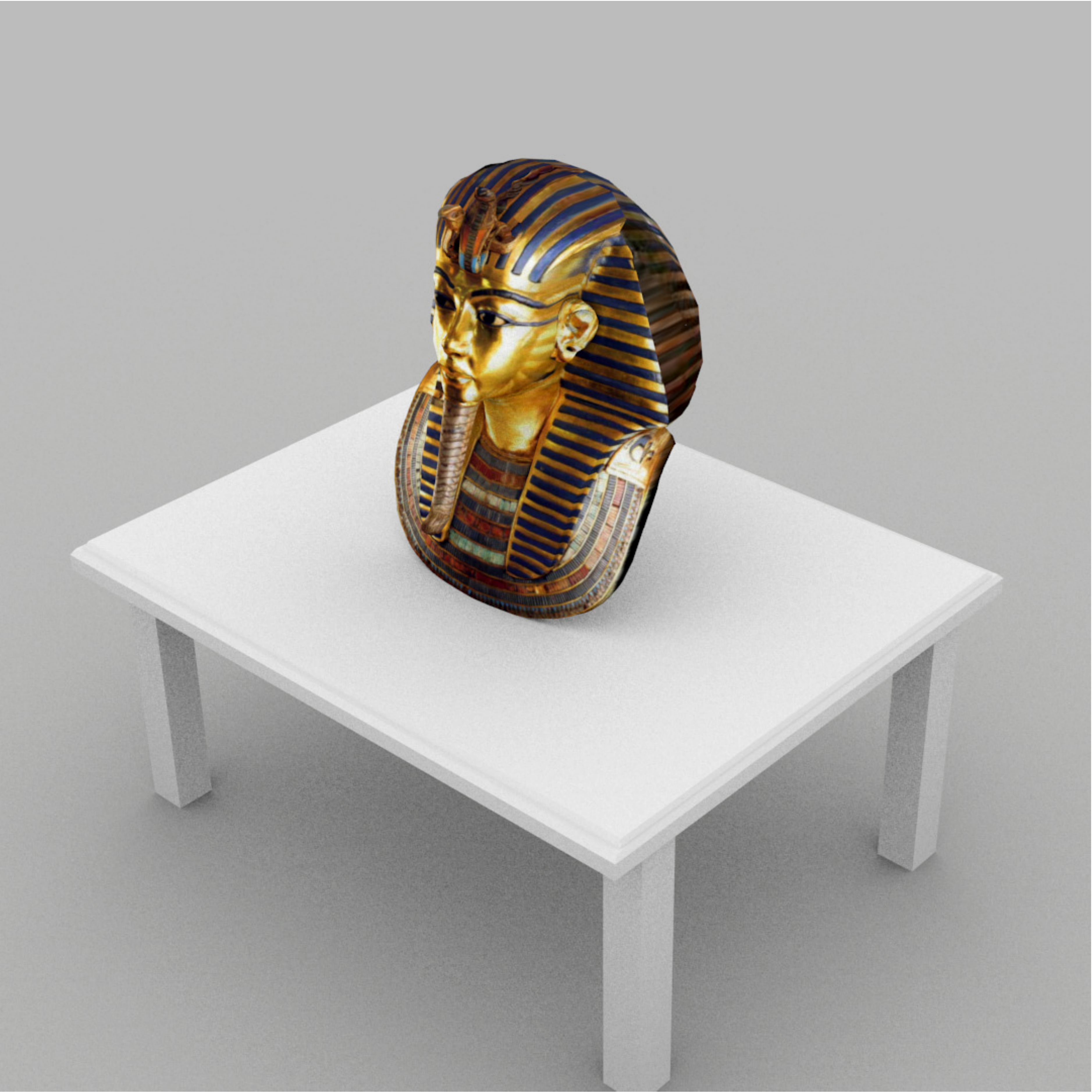}
			\caption{Mask}
			\label{fig:mask}
		\end{subfigure}
		\begin{subfigure}{0.19\linewidth}
			\includegraphics[width=\textwidth]{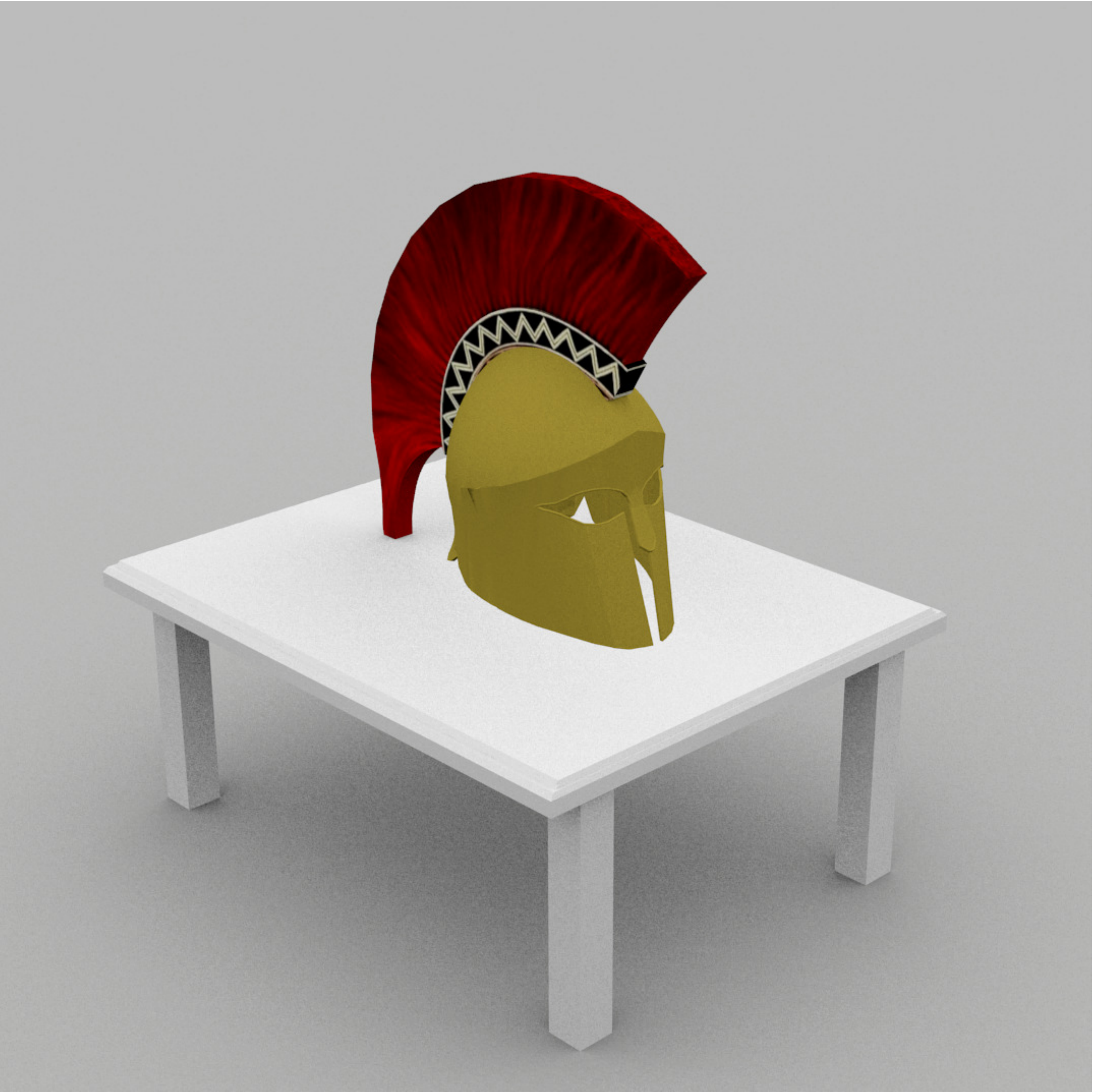}
			\caption{Helmet}
			\label{fig:tiger}
		\end{subfigure}
		\begin{subfigure}{0.19\linewidth}
			\includegraphics[width=\textwidth]{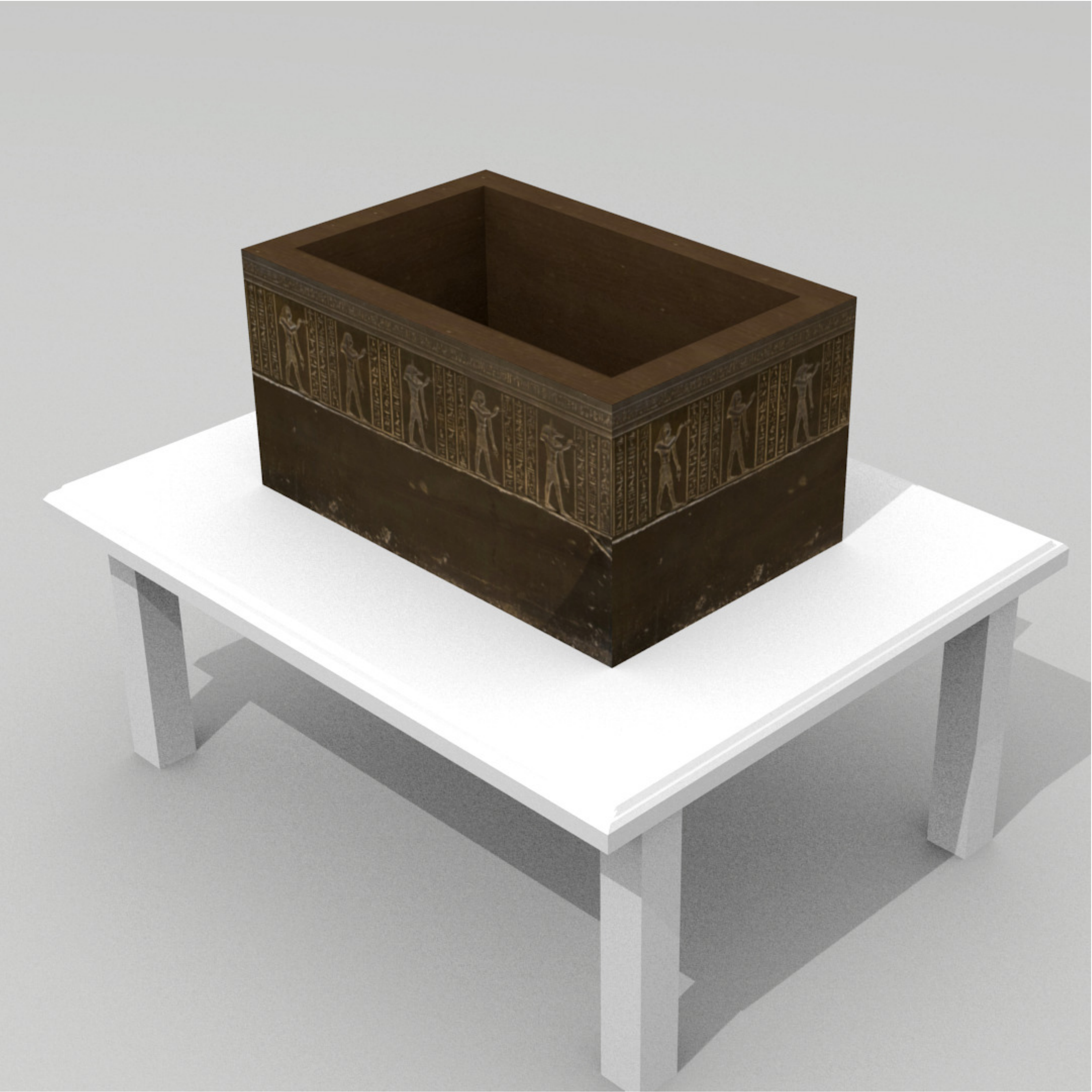}
			\caption{Sarcophagus}
			\label{fig:sarco}
		\end{subfigure}
		\begin{subfigure}{0.19\linewidth}
			\includegraphics[width=\textwidth]{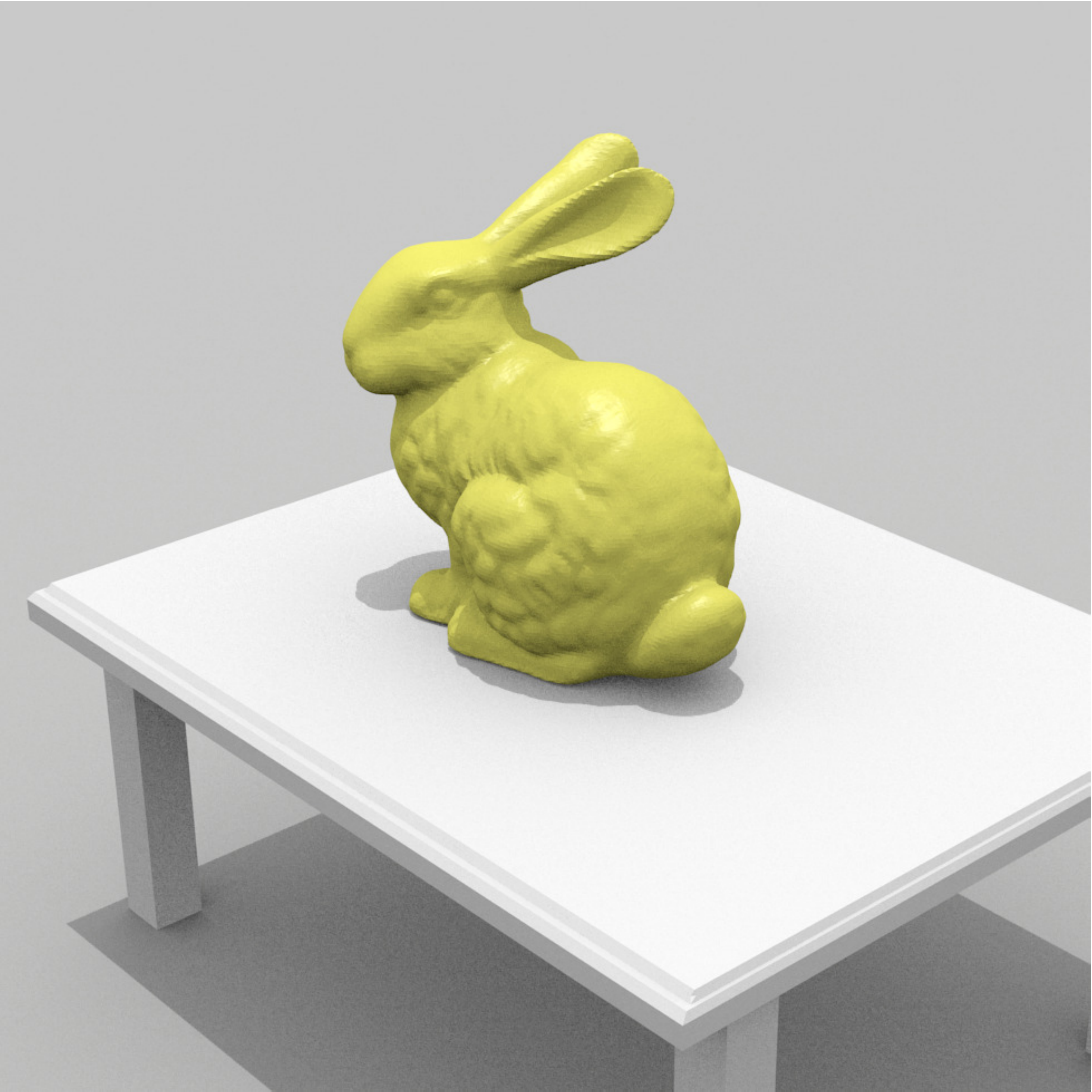}
			\caption{Bunny}
			\label{fig:bunny}
		\end{subfigure}
		\begin{subfigure}{0.19\linewidth}
			\includegraphics[width=\textwidth]{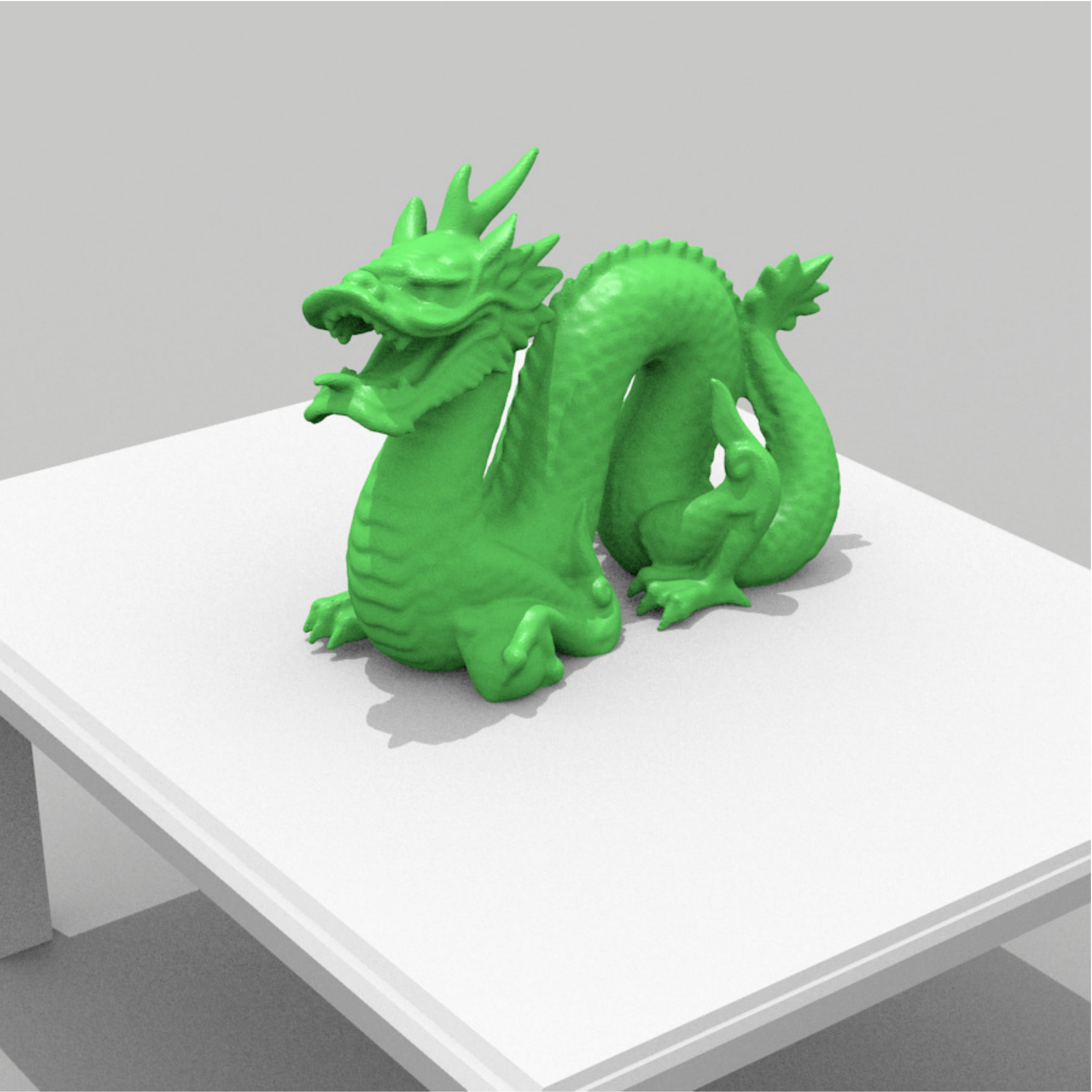}
			\caption{Dragon}
			\label{fig:dragon}
		\end{subfigure}
		\begin{subfigure}{0.19\linewidth}
			\includegraphics[width=\textwidth]{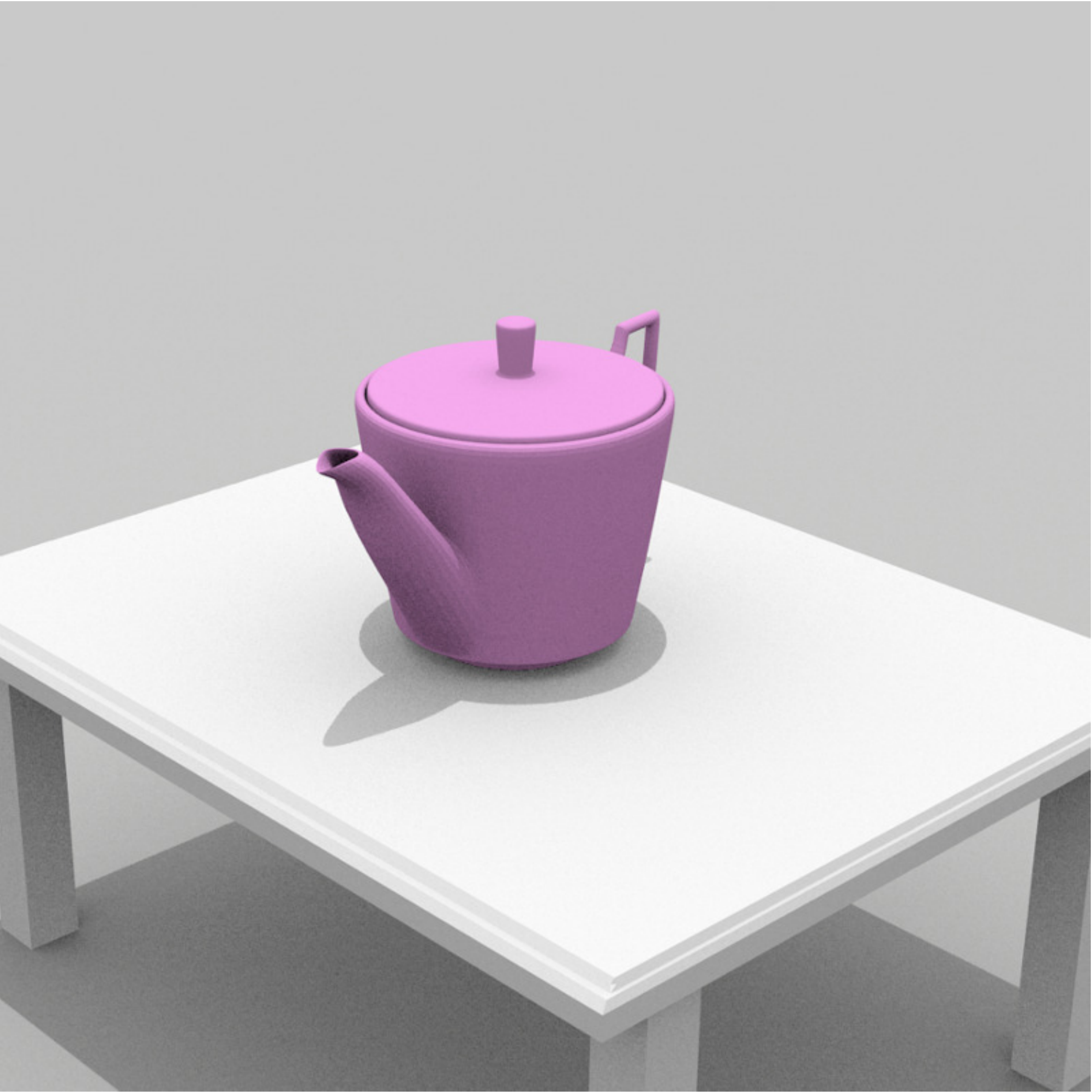}
			\caption{Teapot}
			\label{fig:teapot}
		\end{subfigure}
		\begin{subfigure}{0.19\linewidth}
			\includegraphics[width=\textwidth]{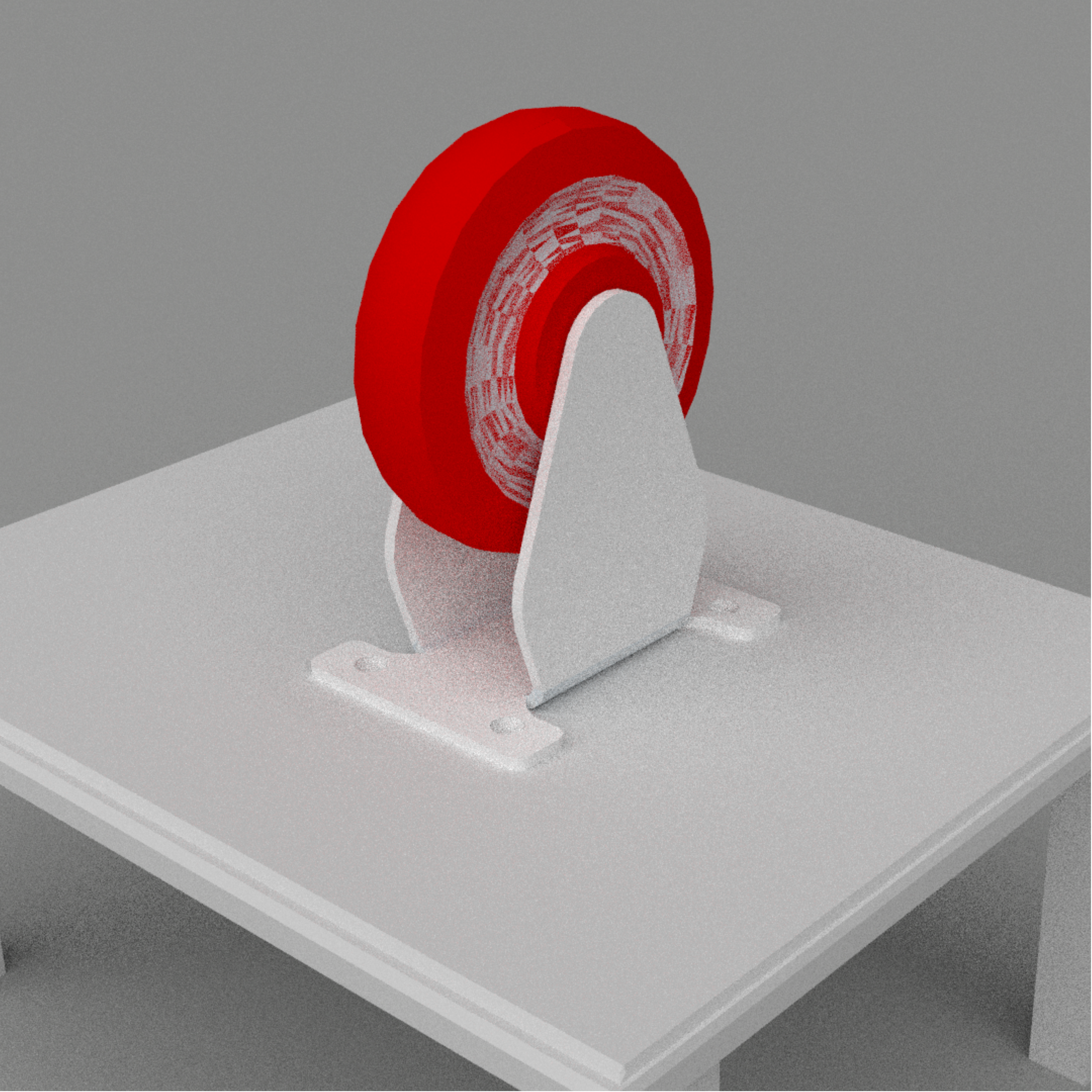}
			\caption{\btxt{Caster wheel}}
			\label{fig:caster}
		\end{subfigure}
		\begin{subfigure}{0.19\linewidth}
			\includegraphics[width=\textwidth]{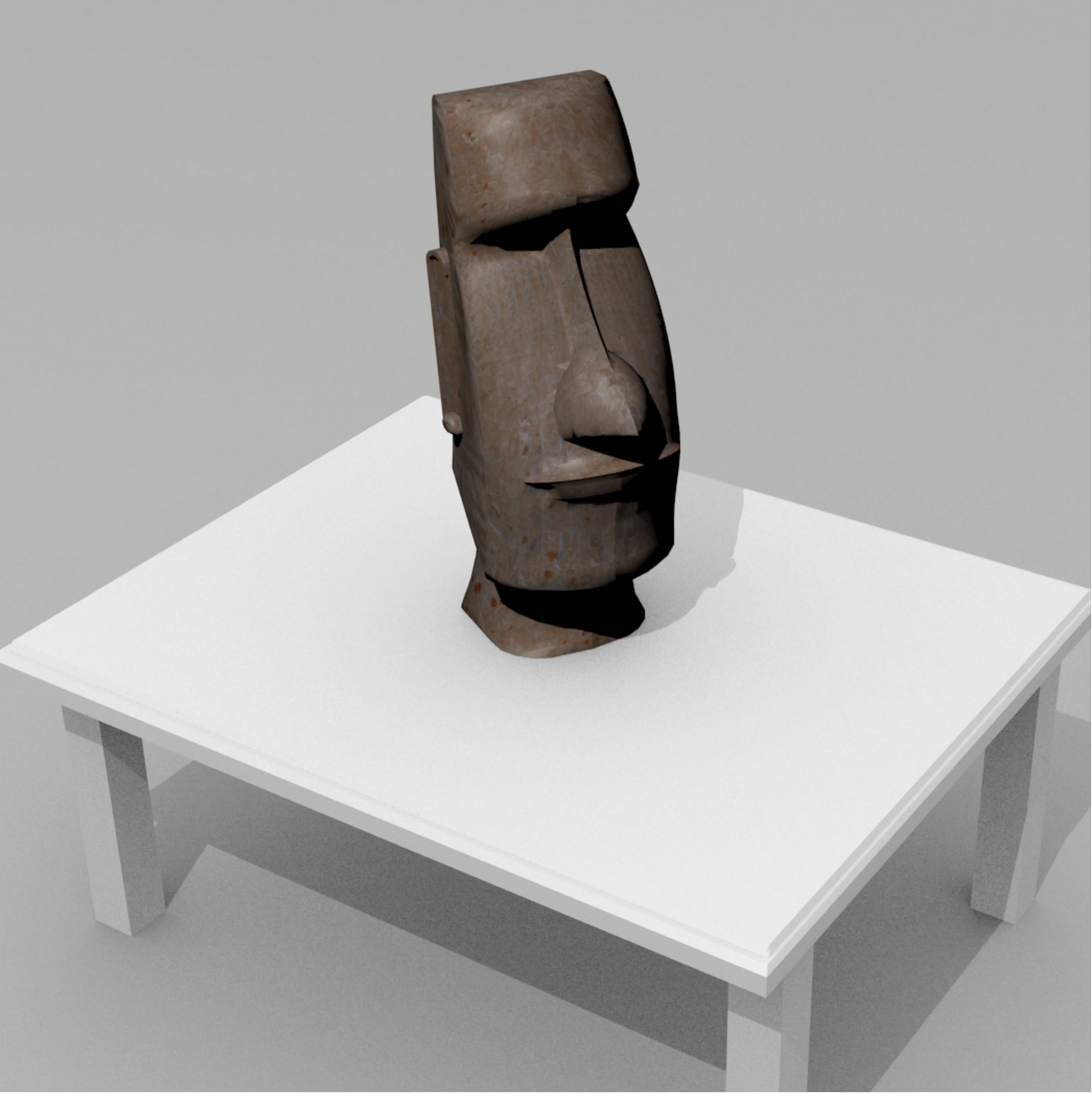}
			\caption{\btxt{Moai head}}
			\label{fig:moai}
		\end{subfigure}
		\begin{subfigure}{0.19\linewidth}
			\includegraphics[width=\textwidth]{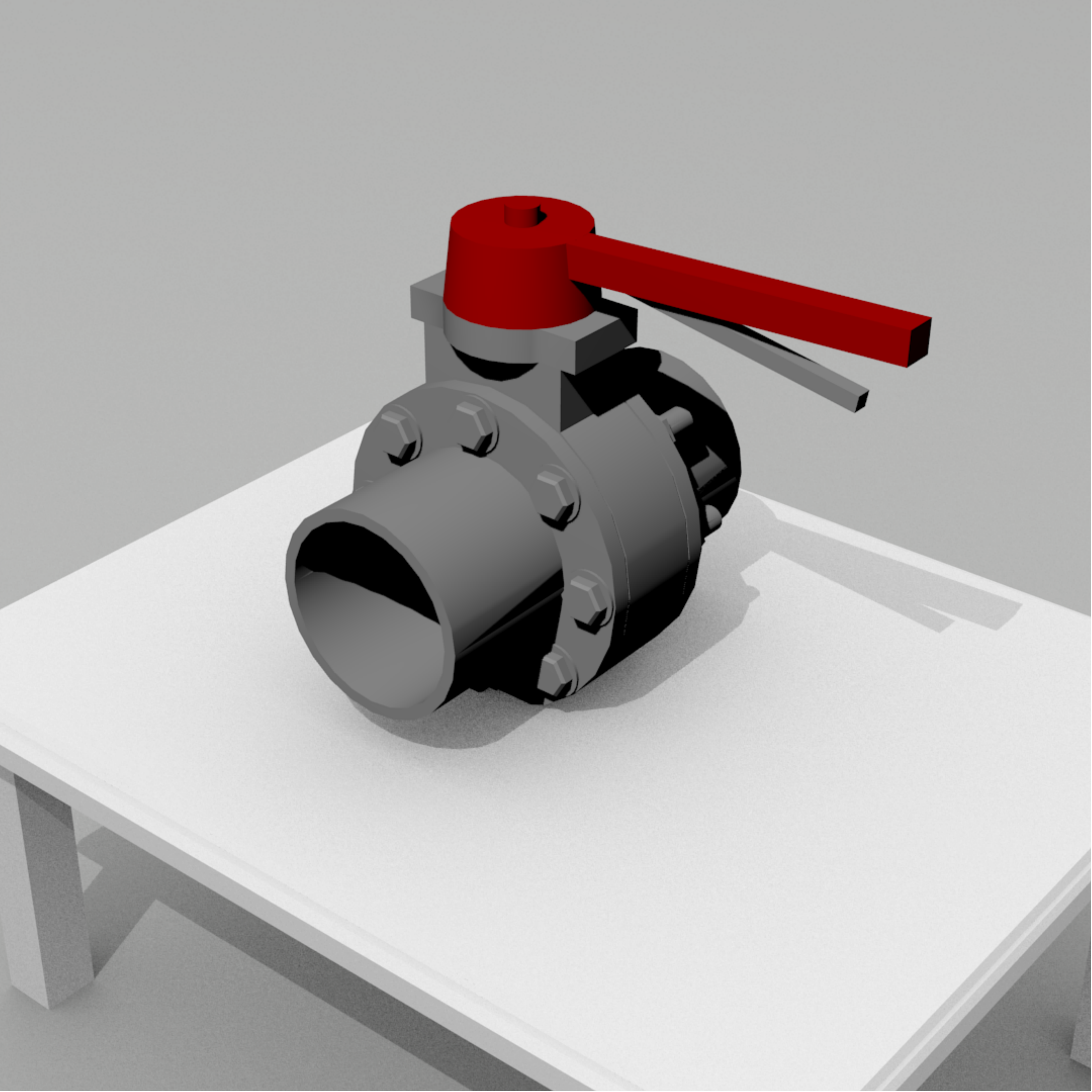}
			\caption{\btxt{Valve}}
			\label{fig:valve}
		\end{subfigure}
		\begin{subfigure}{0.19\linewidth}
			\includegraphics[width=\textwidth]{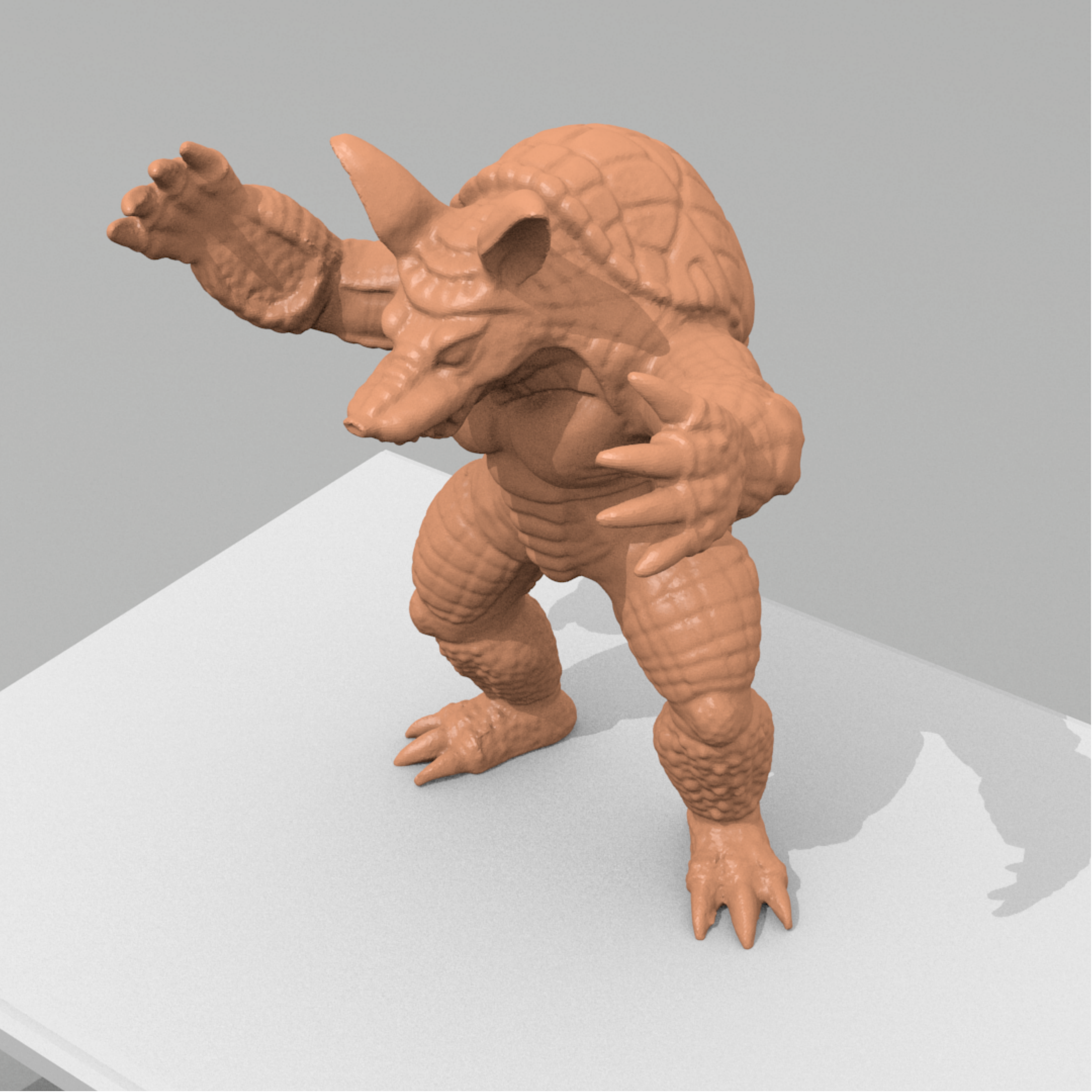}
			\caption{\rtxt{Armadillo}}
			\label{fig:Armadillo}
		\end{subfigure}
			\begin{subfigure}{0.19\linewidth}
		\includegraphics[width=\textwidth]{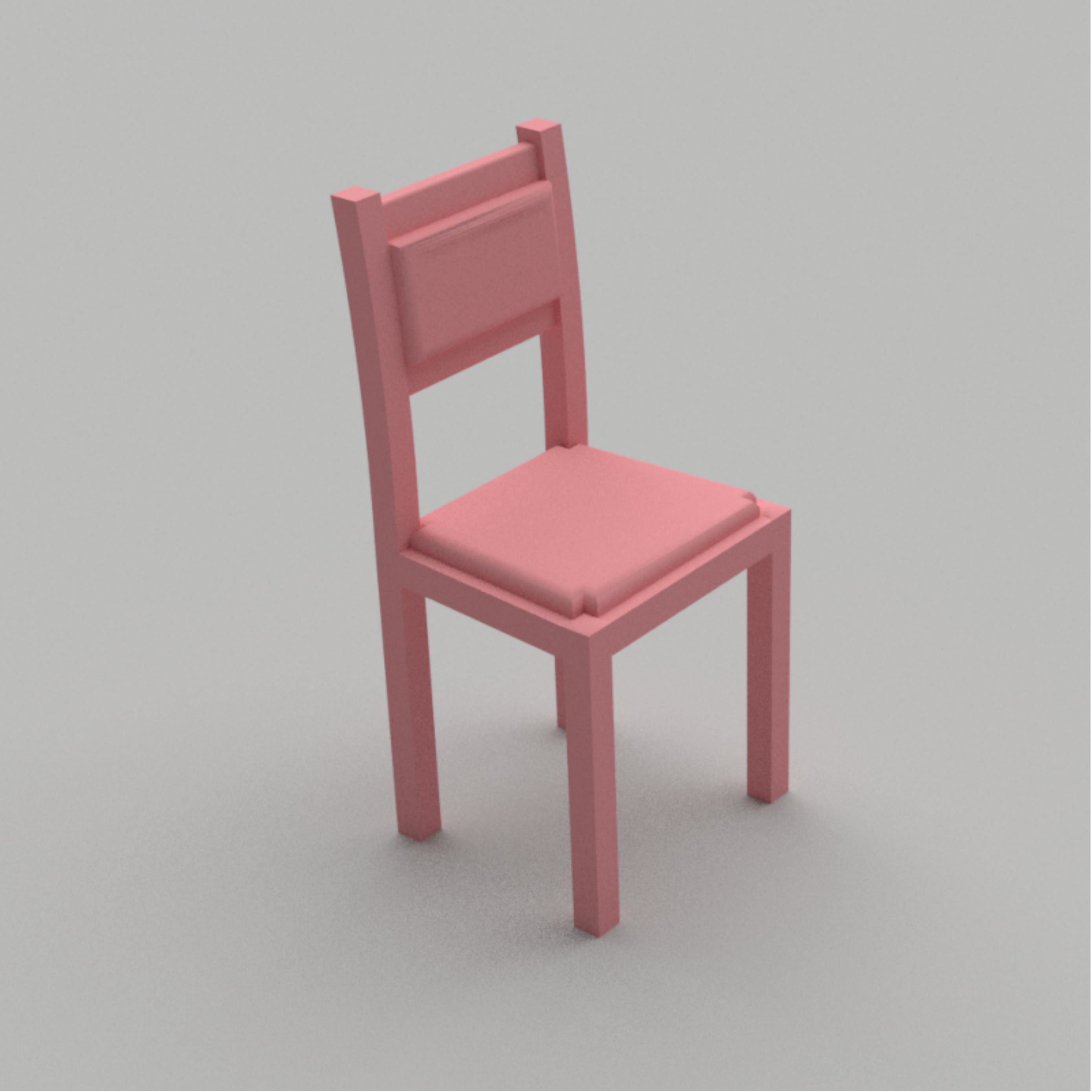}
		\caption{\rtxt{Chair}}
		\label{fig:chair}
		\end{subfigure}
		\begin{subfigure}{0.19\linewidth}
		\includegraphics[width=\textwidth]{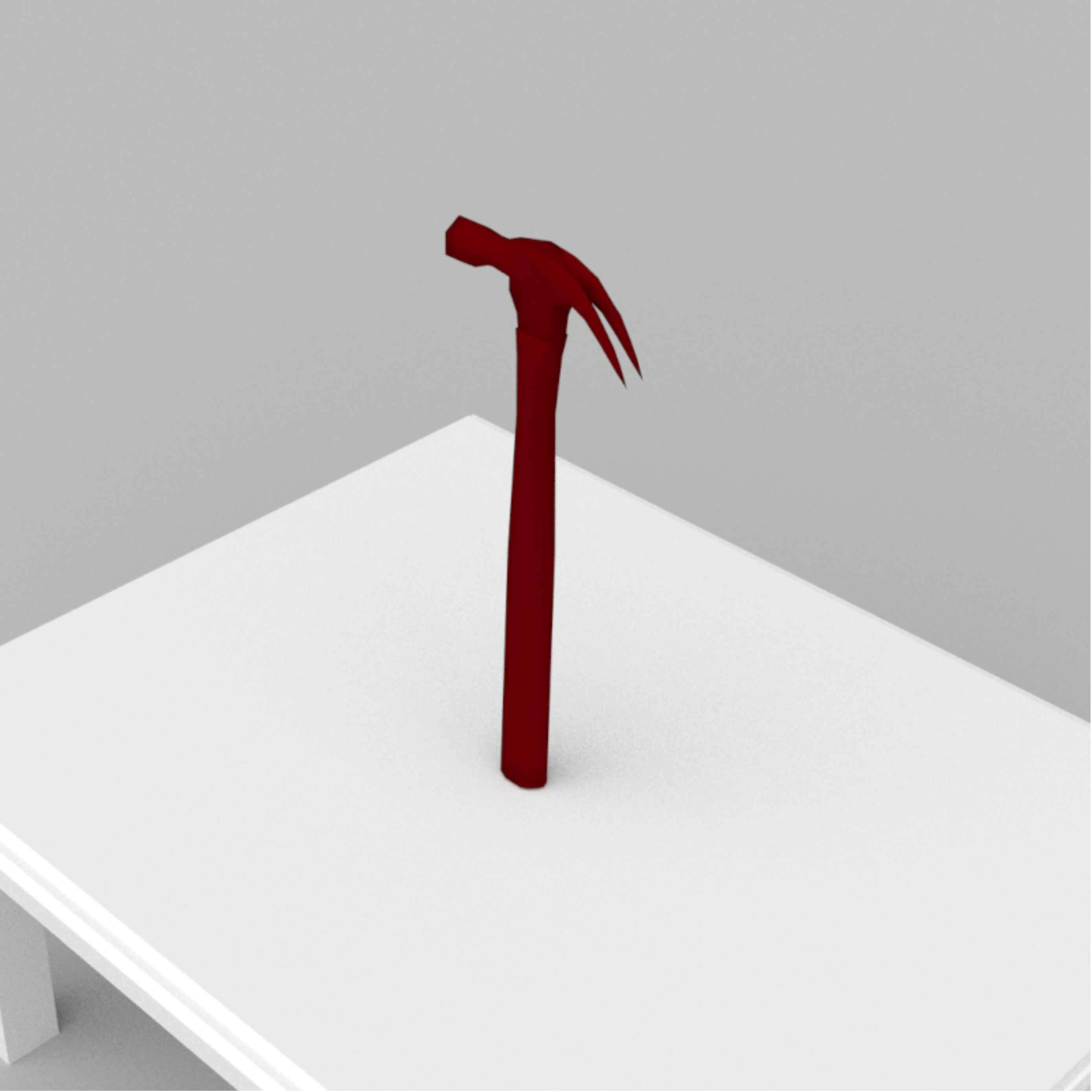}
		\caption{\rtxt{Hammer}}
		\label{fig:Hammer}
		\end{subfigure}
		\caption{3D models used for testing the 3D CNN.}\label{fig:animals}
		\label{fig:reconstruction_objects}
	\end{figure*}

	\subsection{3D Reconstruction of unknown objects}
	
	In this experiment, we will use the the full proposed approach (Fig. \ref{fig:approach}) as the view planner in a simulated 3D reconstruction task, where an unknown object is placed in the scene and the approach has to provide each NBV until the stop criteria is reached. Unlike previous experiment, in this case there is no ground truth NBV since the objects are unknown and they were not used during training, therefore, we will measure the capacity of the method for completing the object. 
	
	The reconstruction scene places the object over a table, \rtxt{except for the chair}. The sensor is simulated with a range camera. The positioning systems is simulated and directly places the sensor in the planned pose. The performance metric is the percentage of coverage, calculated as the ratio between the matching scanned points and the points in the reference model given a distance of 0.001m. Scale factor $k$ was set to 2.5. The objects to be modeled are \rtxt{thirteen}: a sphere, the mask of Tutankhamun, a Corinthian helmet, a Egyptian sarcophagus, the Stanford bunny, the Stanford dragon, a teapot, a caster wheel, a Moai head, a butterfly valve, \rtxt{an armadillo, a chair and a hammer}. The objects are depicted in Fig. \ref{fig:reconstruction_objects}. Our simulation was implemented using the view planning library (VPL) \cite{vasquez2017view} and Blensor simulator \cite{gschwandtner2011blensor}.

	\subsubsection{Generalization}

	First, we test the different network variations on three unseen objects that were not used during training nor validation; this was done to show which architecture provides the best generalization despite training and validation performance. For this purpose, we have selected the sphere, the bunny and the dragon, which contrast to dataset objects because of either their convex or elongated shapes. For all the variants, the initial sensor location is placed in front of the object. Once the first scan is integrated to the probabilistic grid, each network makes its prediction and the reconstruction continues, in consequence, each variation follows a different sequence of sensing locations. The experiment was done for ten scans in order to observe the coverage reached for each variant.

	\begin{figure}[tb]
		\centering
		\includegraphics[width=\linewidth]{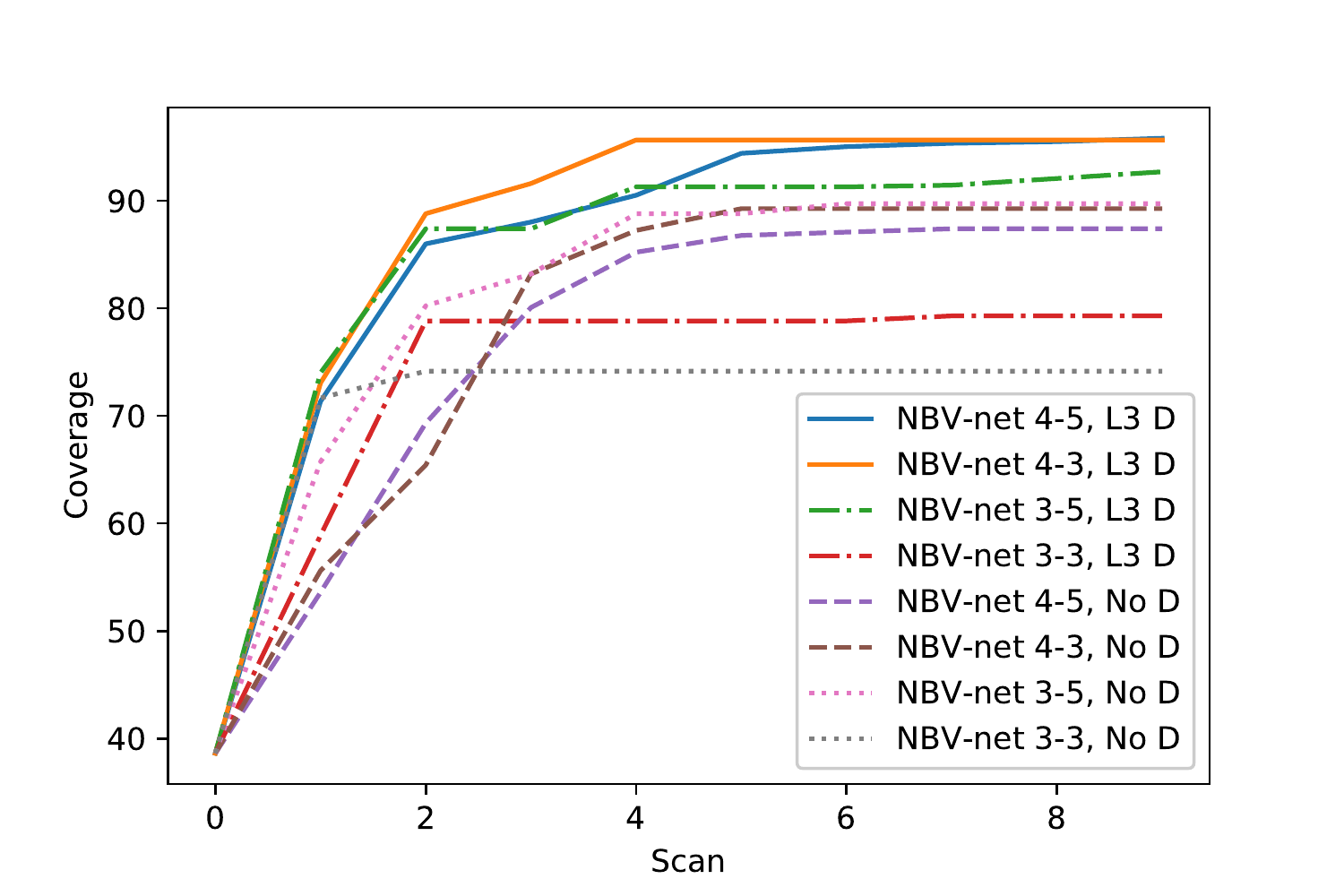}
		\caption{Reconstruction coverage for the sphere object.}
		\label{fig:coverage_sphere}
	\end{figure}
	
	\begin{figure}[tb]
		\centering
		\includegraphics[width=\linewidth]{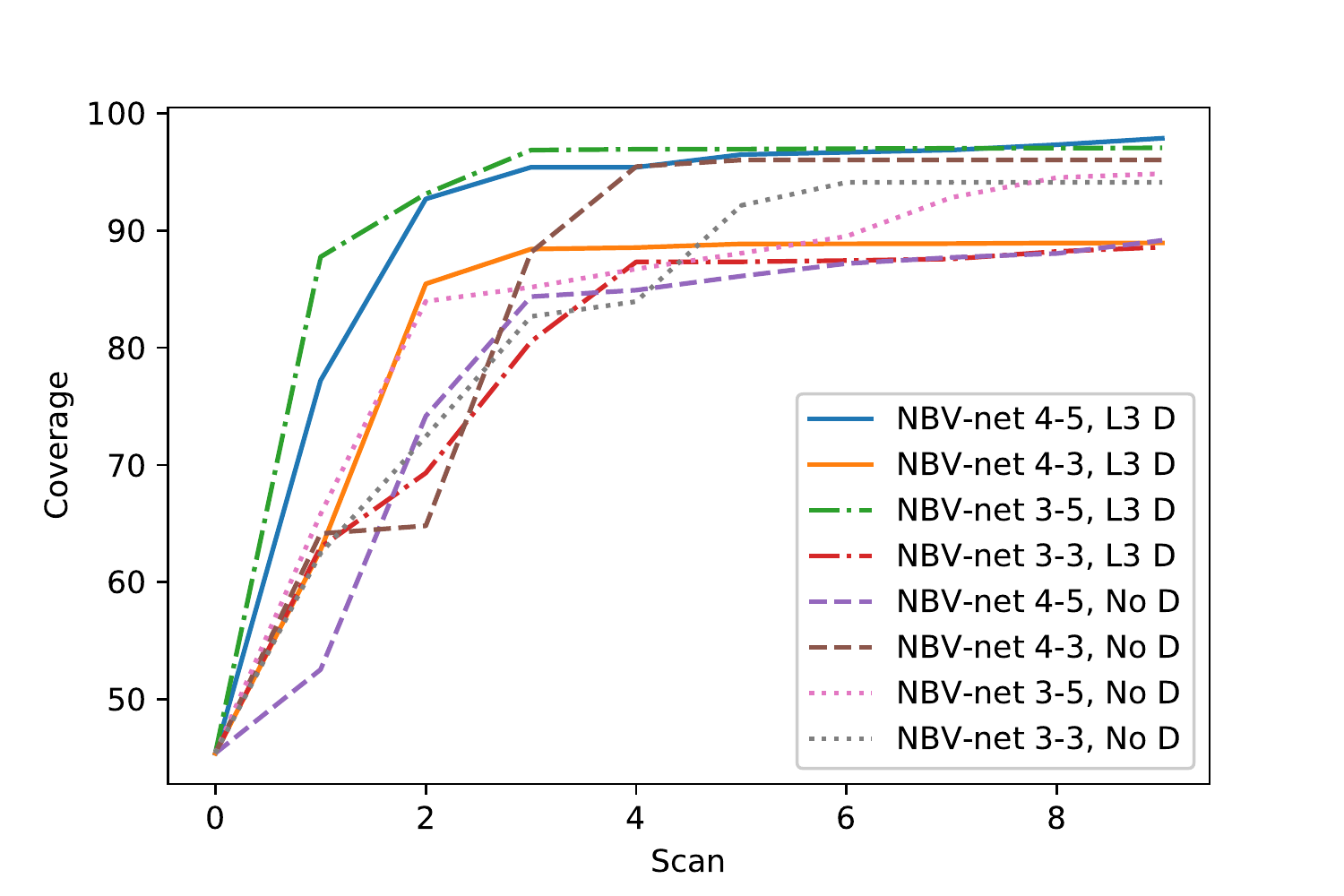}
		\caption{Reconstruction coverage for the bunny object.}
		\label{fig:coverage_bunny}
	\end{figure}

	\begin{figure}[tb]
		\centering
		\includegraphics[width=\linewidth]{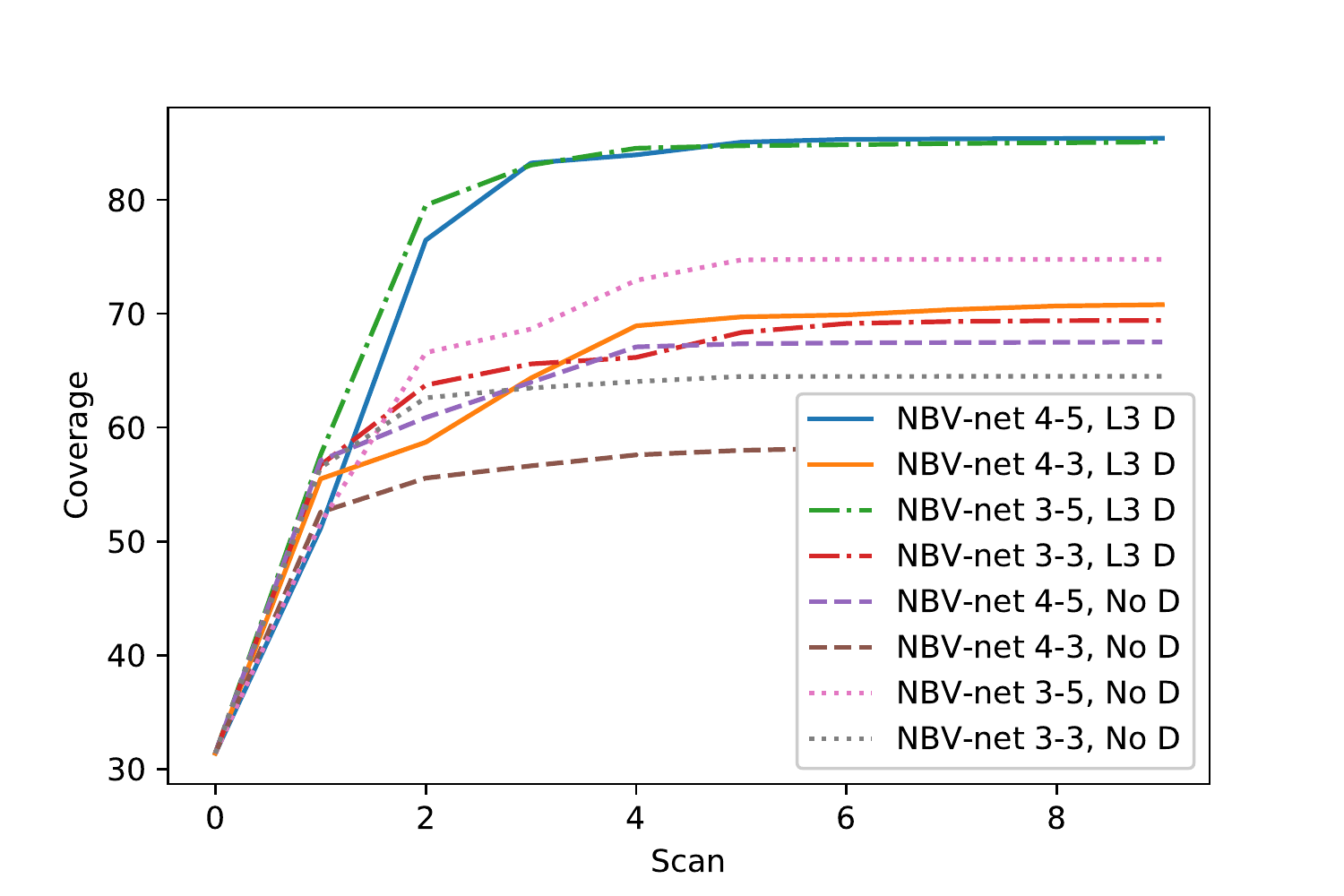}
		\caption{Reconstruction coverage for the dragon object.}
		\label{fig:coverage_dragon}
	\end{figure}

	As we can see in Figs. \ref{fig:coverage_sphere}, \ref{fig:coverage_bunny} and \ref{fig:coverage_dragon}, the architecture that has the largest number of convolutional and fully connected layers (NBV-net 4-5) with dropout from convolutional layer 3, was the one with the best performance, that is, the largest final reconstruction percentage. Its reconstruction coverage ranged from around 85\% up to 95\%. We believe that this result is due to the increment in the network’s feature extraction layers (four wider convolutional layers), also due to the dropout that avoided overfitting over the training set, and due to the five fully connected layers that enhanced the network’s capabilities to perform regression. It is worth to say that network NBV-net 3-5 with dropout, had a tendency to cover greater portions of the objects at the beginning of the scanning. A possible explanation for this phenomenon is that with fewer convolutional layers less features were modeled, but those were important to identify poses that provide coverage of large portions of the object; however, once the scanning progresses, the missing modeled features are important to cover details of the object, what NBV-net 4-5 was indeed able to do.

	\subsubsection{Different object shapes}
	
	We carry out a comparison, in terms of percentage of object reconstruction and processing time needed to compute the next-best-view, between the method proposed in this work (NBV-net 4-5) with other two related methods.
	In \cite{mendoza2018paper}, the authors address the next-best-view problem with a classification-based approach.
	The output of the 3D-CNN is limited to a set of 14 possible sensor views. 
	The approach in \cite{kriegel2012next} proposed an information gain based method that exhaustively evaluates 20 views around the object.
	\textcolor{black}{The three approaches were tested in the
		reconstruction of \rtxt{thirteen} proposed unknown objects (Fig. \ref{fig:reconstruction_objects}). The stop criteria was ten scans.}
	
	\begin{table}
		\centering
		\caption{Reconstruction coverage for each tested object.}
		\begin{tabular}{|c|c|c|c|}
			\hline 
			Object & Classif. & Regression & Inf. Gain \\  
			\hline 
			Sphere & \textbf{96.7} & 95.7 & 96.2 \\
			Mask  & 94.73 & \textbf{95.32} & 95.0 \\ 
			Helmet & 82.7 & 84.9 & \textbf{86.5} \\ 
			Sarcophagus & 64.2 & 71.7 & \textbf{94.1} \\ 
			Bunny  & 90.0 & 97.8 & \textbf{98.1} \\
			Dragon & 71.3 & 85.4 & \textbf{90.4} \\
			Teapot & 87.1 & 93.0 & \textbf{93.2} \\
			\textcolor{black}{Caster} & 89.1 & 90.6 & \textbf{100}  \\
			\textcolor{black}{Moai Head} & 96.8 & 97.1 & \textbf{98.9} \\
			\textcolor{black}{Valve} & 73.3 & 70.7 & \textbf{85.9} \\ 
			\rtxt{Armadillo} &  84.6 & 86.0 & \textbf{95.2} \\
			\rtxt{Chair} & 85.2 & 84.6 & \textbf{89.4} \\
			\rtxt{Hammer} & 53.2 & 56.8 & \textbf{57.0} \\
			\hline 
		\end{tabular} 
		\label{tab:coverage}
	\end{table}
	\begin{table}
		\centering
		\caption{Processing time for next-best-view computation}
		\begin{tabular}{|c|c|c|c|}
			\hline 
			& Classif.  & Regression & Inf. Gain \\ 
			\hline 
			Time & \textbf{0.01 s} & 0.3 s & 29.9 s \\ 
			\hline 
		\end{tabular} 
		\label{tab:timem}
	\end{table}

	%Unlike the previous experiment where the networks were left running until they reached ten scans, we apply a common stop criterion, to stop once the planned view does not increase the scanned surface.

	\begin{figure*}[tb]
		\centering
		\begin{subfigure}{0.19\linewidth}
			\includegraphics[width=\textwidth]{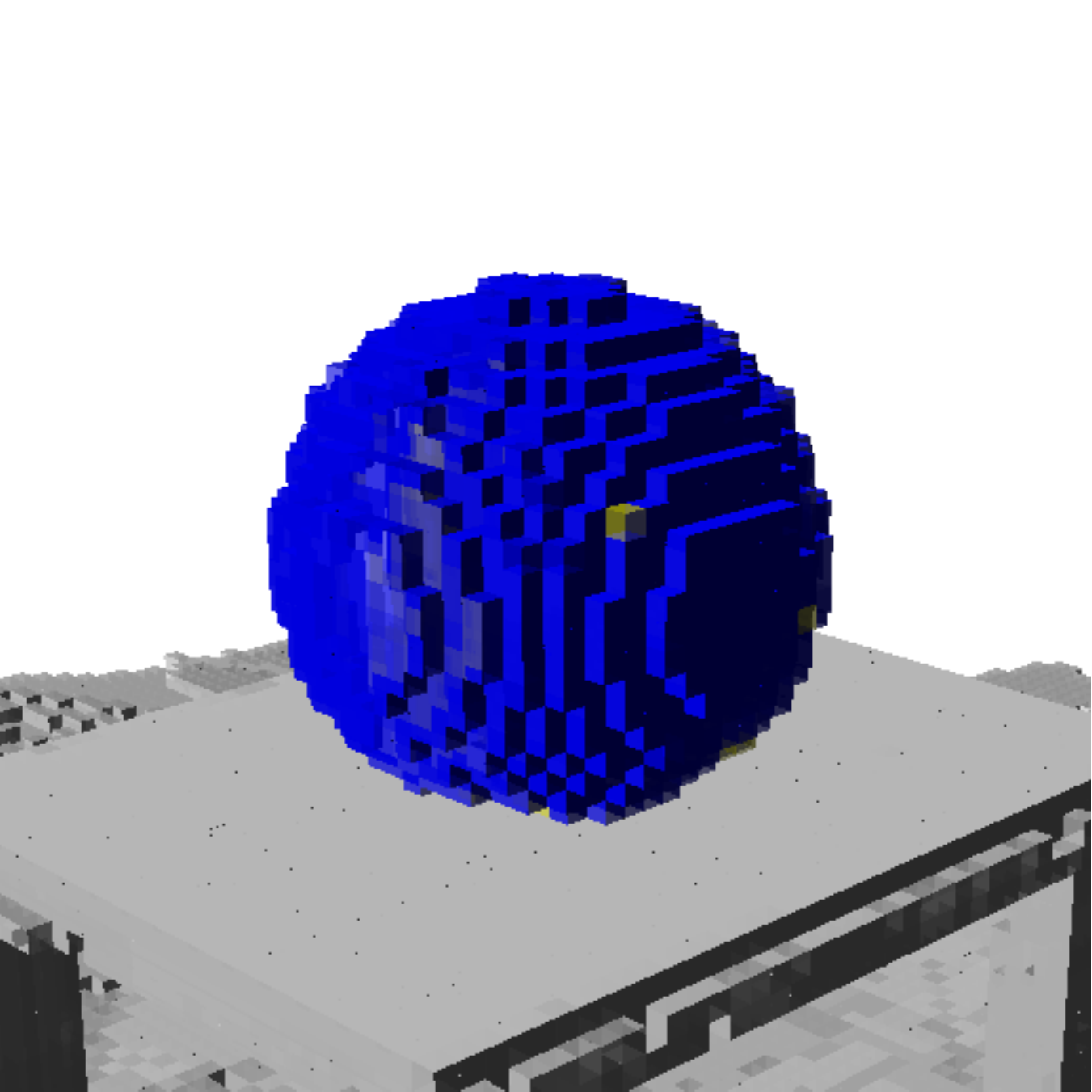}
			\caption{Sphere}
			\label{fig:rsphere}
		\end{subfigure}
		\begin{subfigure}{0.19\linewidth}
			\includegraphics[width=\textwidth]{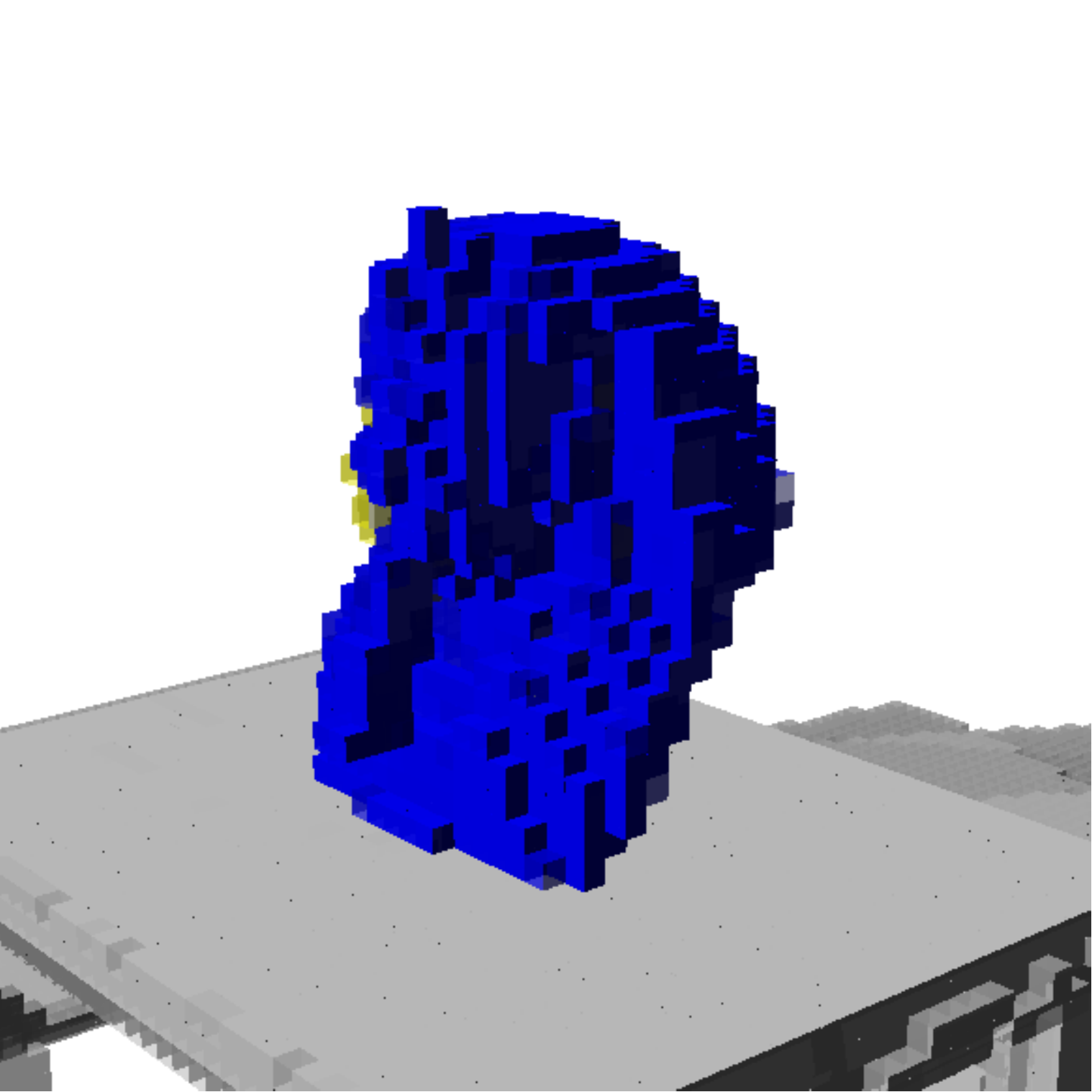}
			\caption{Mask}
			\label{fig:rmask}
		\end{subfigure}
		\begin{subfigure}{0.19\linewidth}
			\includegraphics[width=\textwidth]{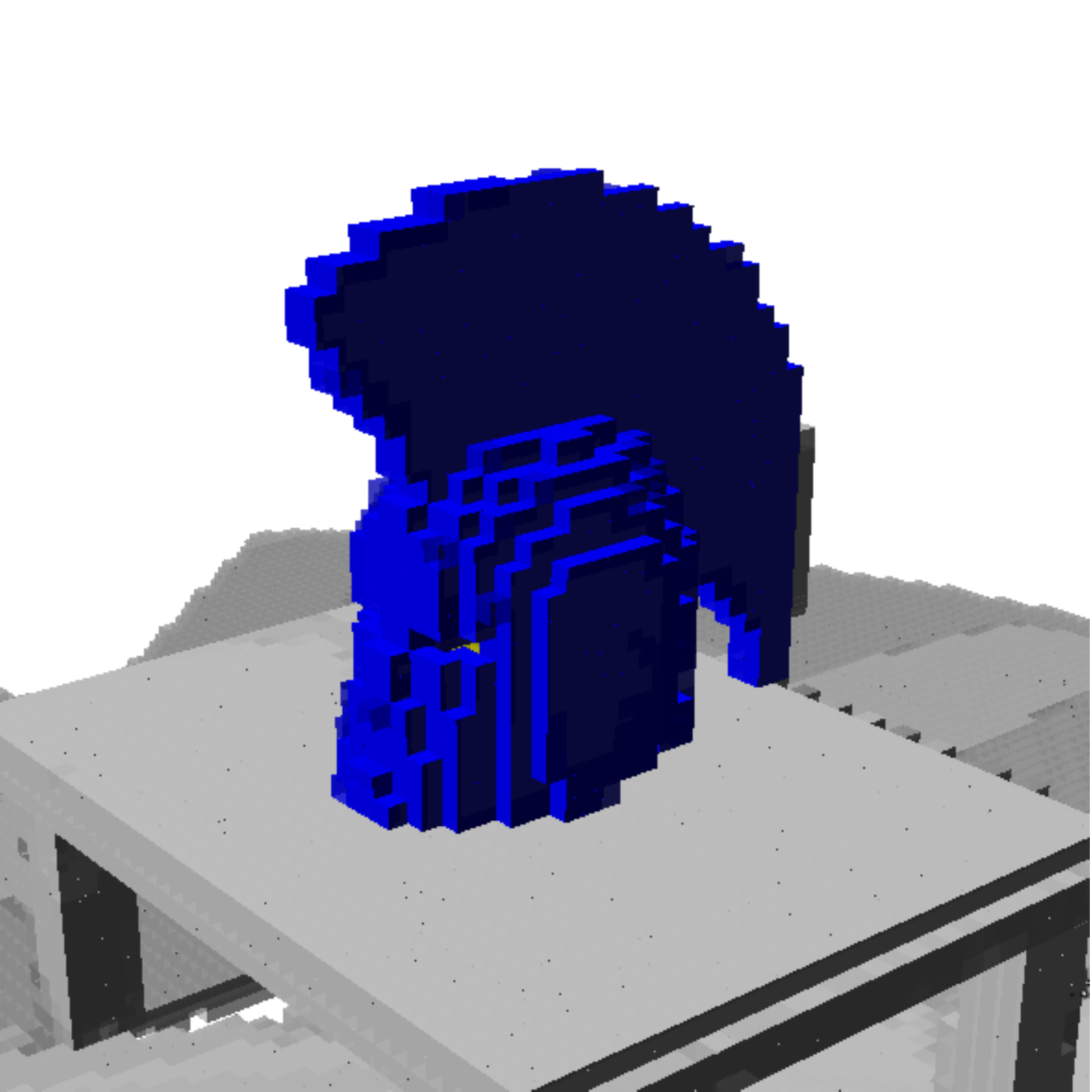}
			\caption{Helmet}
			\label{fig:rtiger}
		\end{subfigure}
		\begin{subfigure}{0.19\linewidth}
			\includegraphics[width=\textwidth]{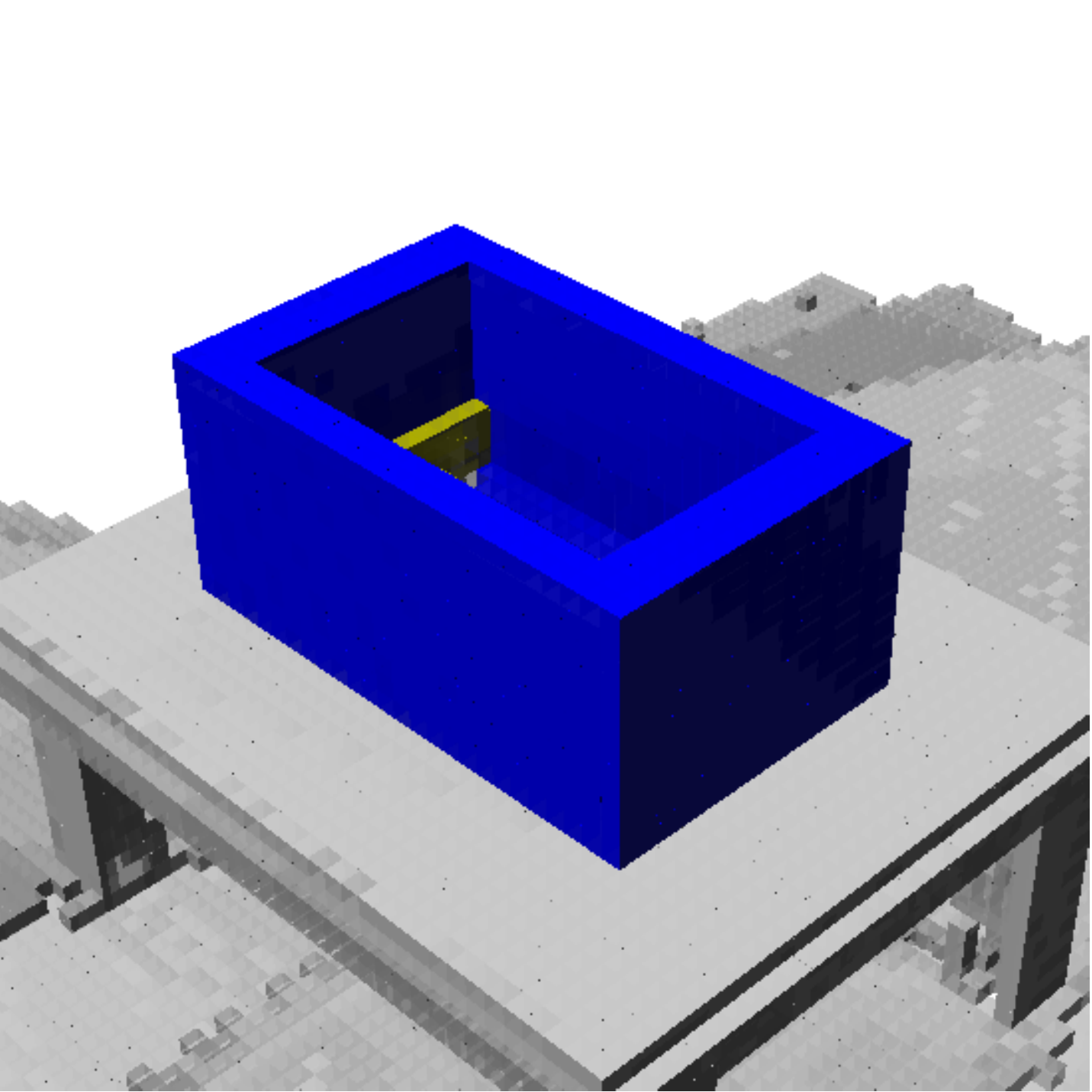}
			\caption{Sarcophagus}
			\label{fig:rsarco}
		\end{subfigure}
		\begin{subfigure}{0.19\linewidth}
			\includegraphics[width=\textwidth]{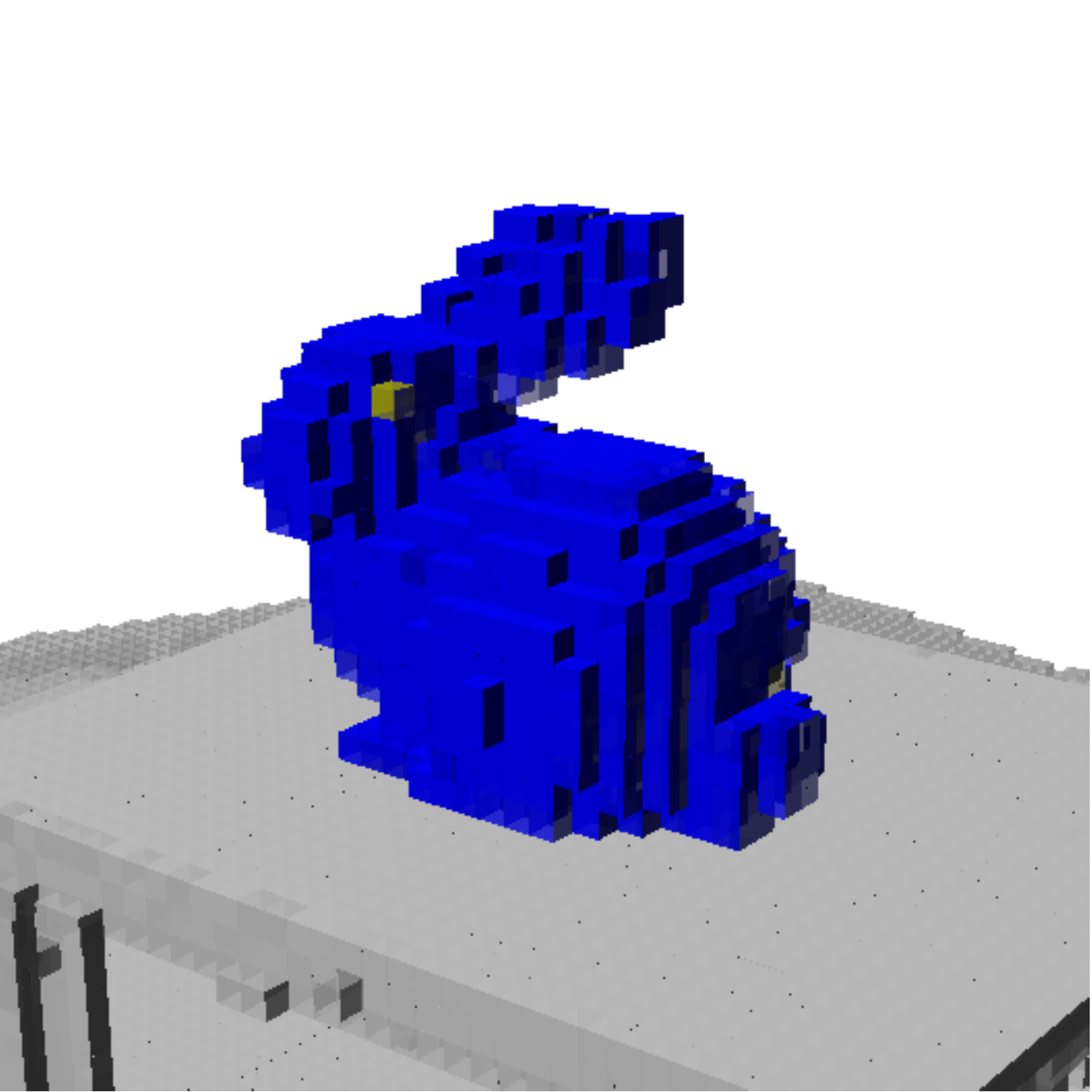}
			\caption{Bunny}
			\label{fig:rbunny}
		\end{subfigure}
		\begin{subfigure}{0.19\linewidth}
			\includegraphics[width=\textwidth]{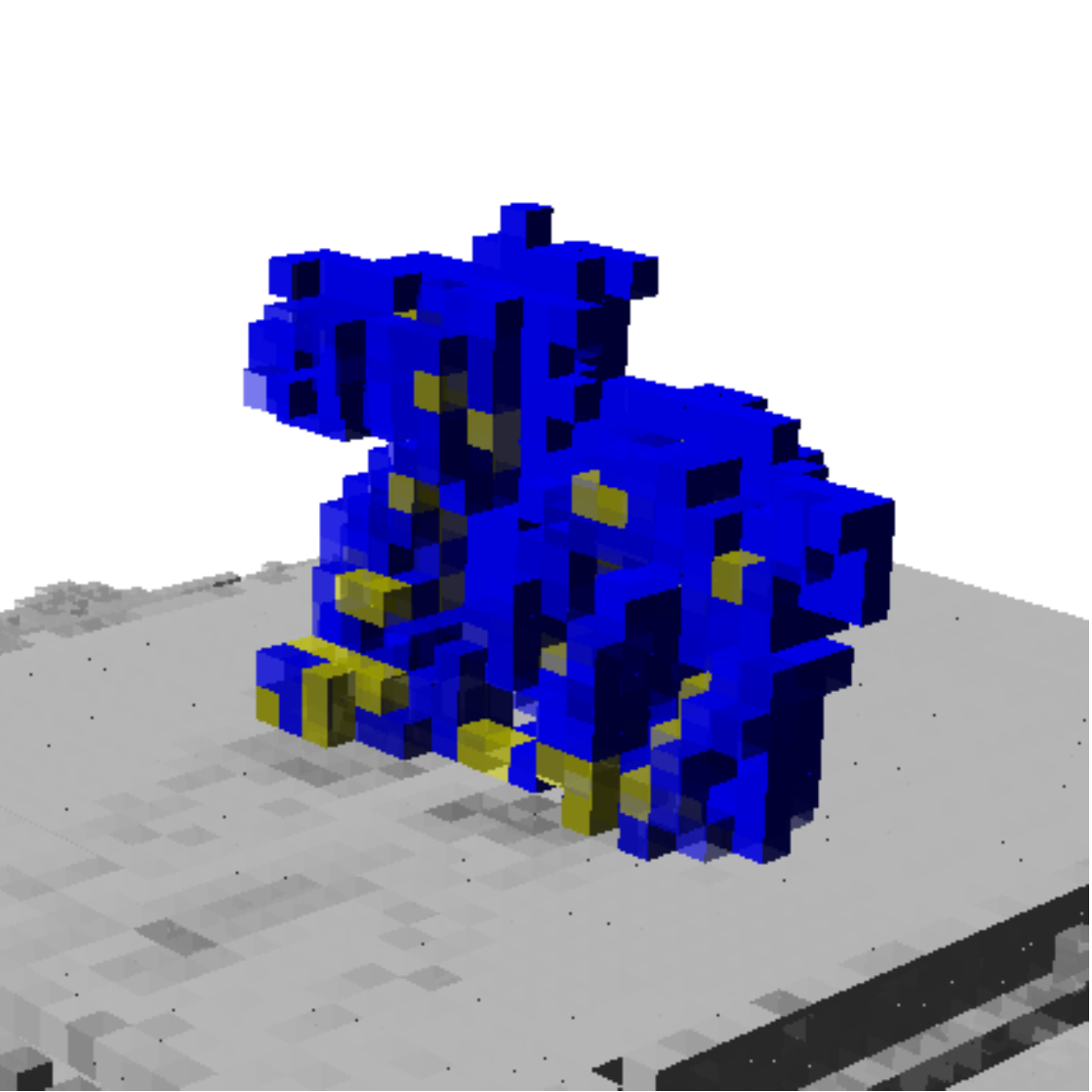}
			\caption{Dragon}
			\label{fig:rdragon}
		\end{subfigure}
		\begin{subfigure}{0.19\linewidth}
			\includegraphics[width=\textwidth]{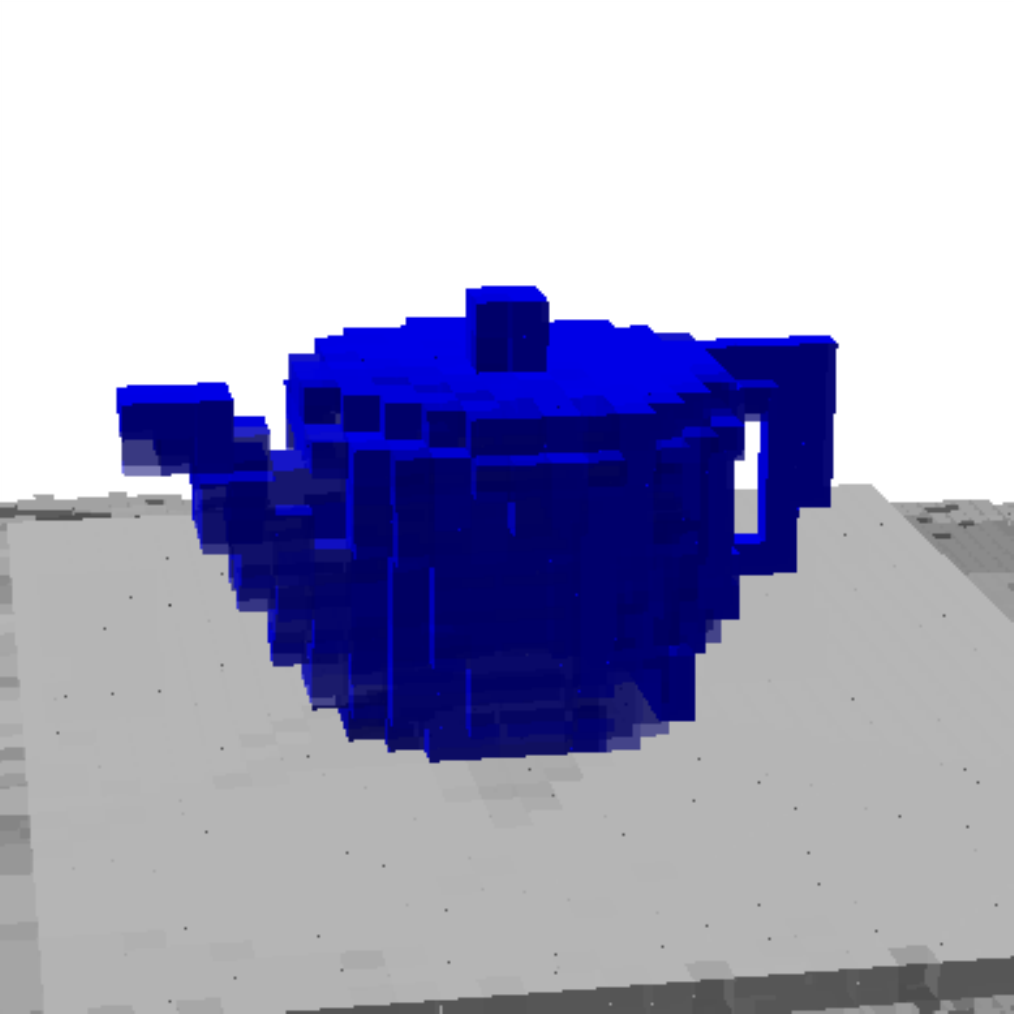}
			\caption{Teapot}
			\label{fig:rteapot}
		\end{subfigure}
		\begin{subfigure}{0.19\linewidth}
			\includegraphics[width=\textwidth]{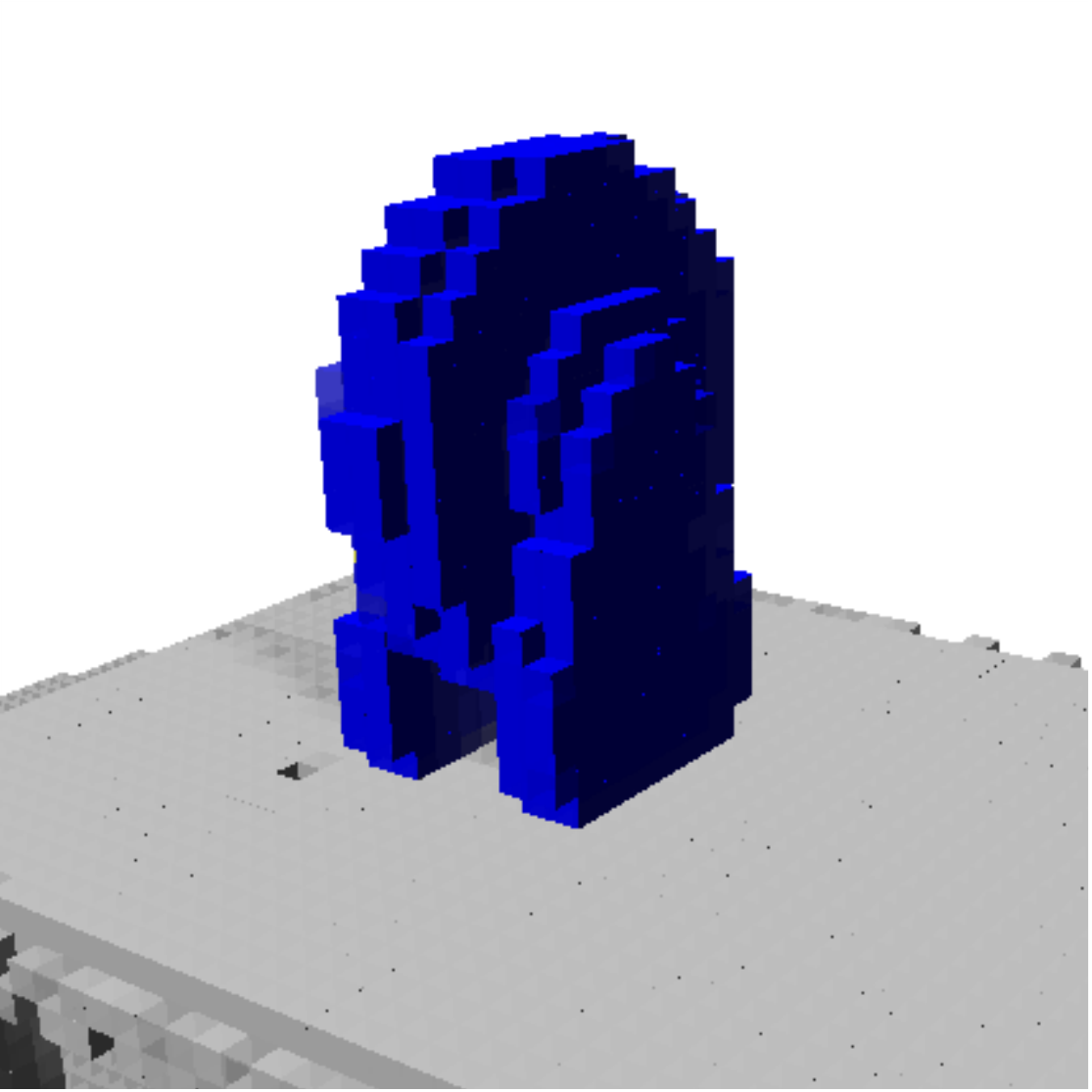}
			\caption{\btxt{Caster wheel}}
			\label{fig:rcaster}
		\end{subfigure}
		\begin{subfigure}{0.19\linewidth}
			\includegraphics[width=\textwidth]{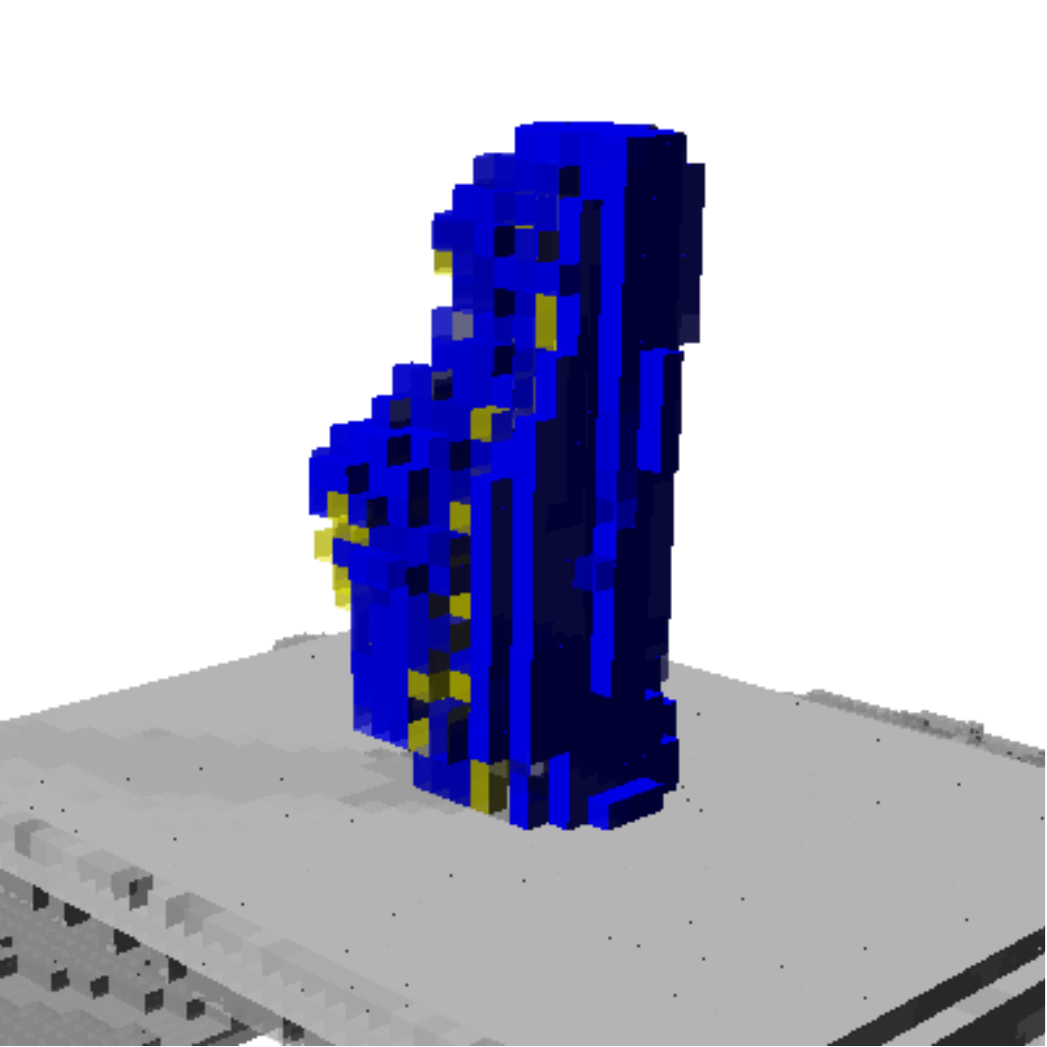}
			\caption{\btxt{Moai head}}
			\label{fig:rmoai}
		\end{subfigure}
		\begin{subfigure}{0.19\linewidth}
			\includegraphics[width=\textwidth]{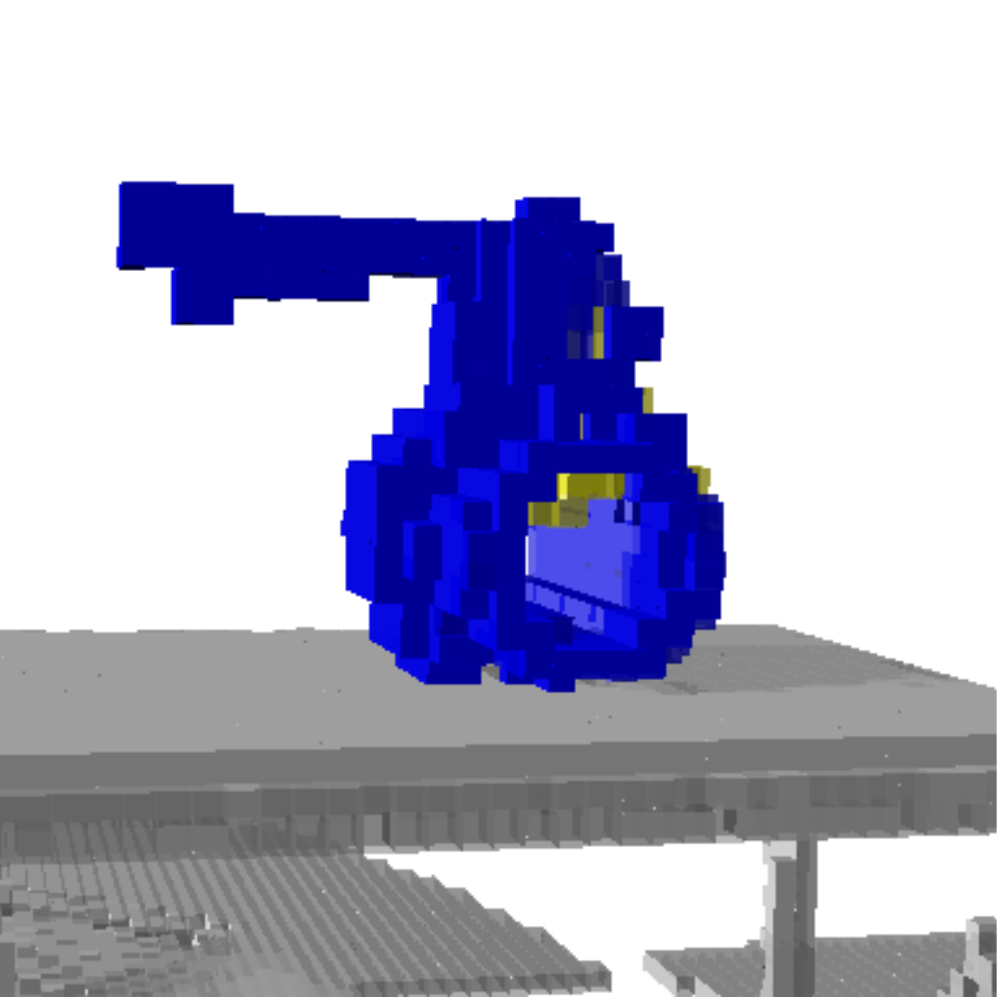}
			\caption{\btxt{Valve}}
			\label{fig:rvalve}
		\end{subfigure}
		\begin{subfigure}{0.19\linewidth}
			\includegraphics[width=\textwidth]{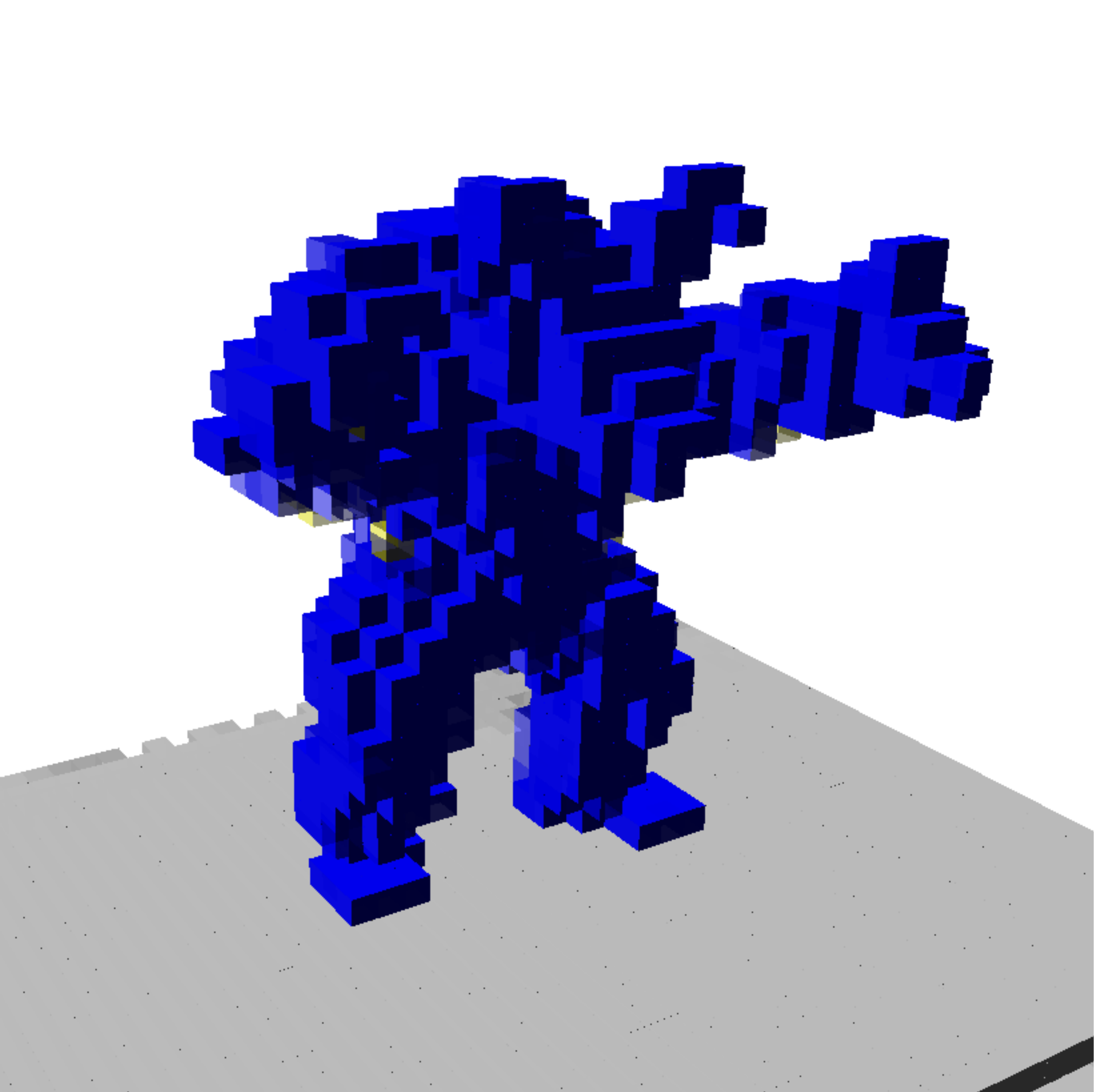}
			\caption{\rtxt{Armadillo}}
			\label{fig:hammer}
		\end{subfigure}
		\begin{subfigure}{0.19\linewidth}
			\includegraphics[width=\textwidth]{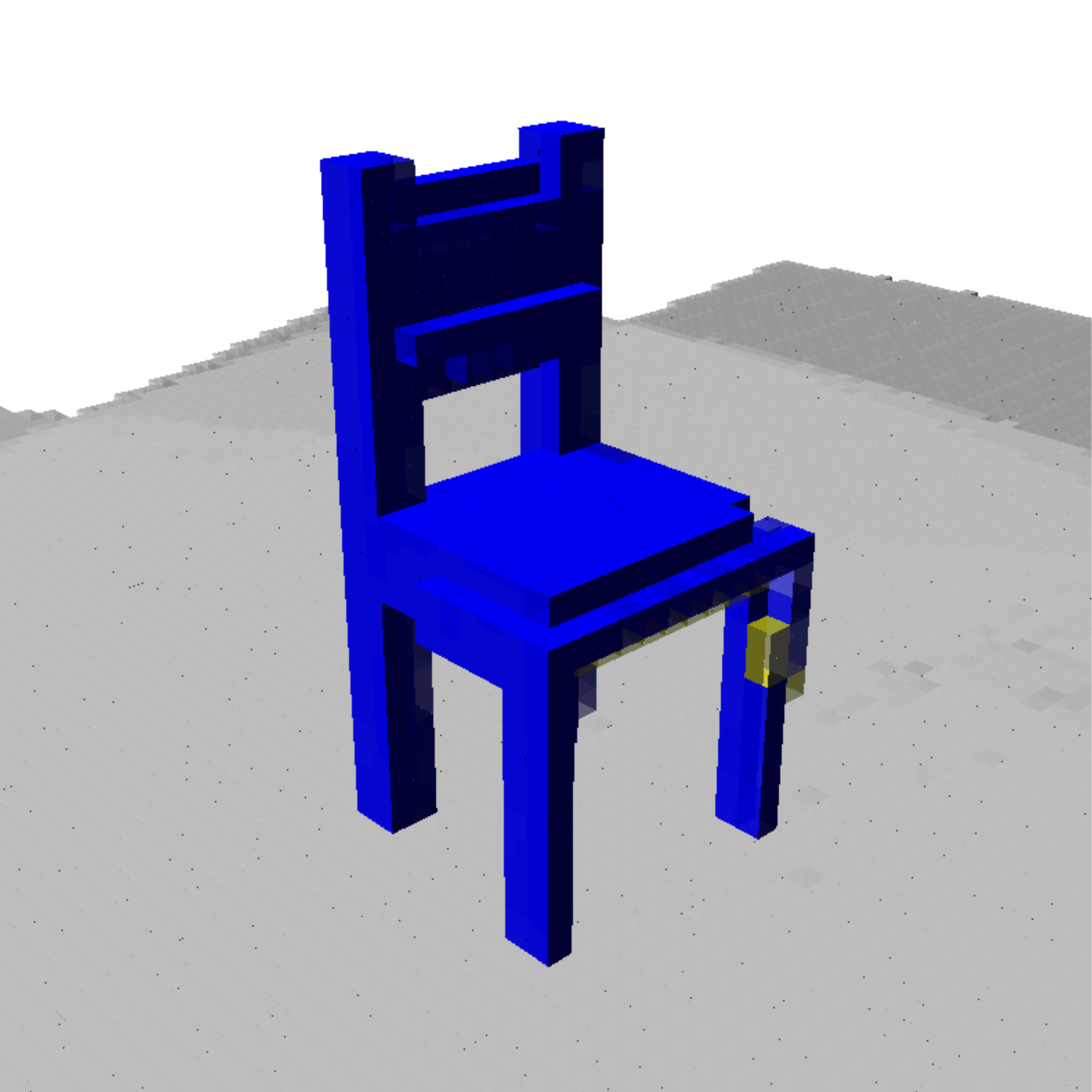}
			\caption{\rtxt{Chair}}
			\label{fig:hammer}
		\end{subfigure}
		\begin{subfigure}{0.19\linewidth}
			\includegraphics[width=\textwidth]{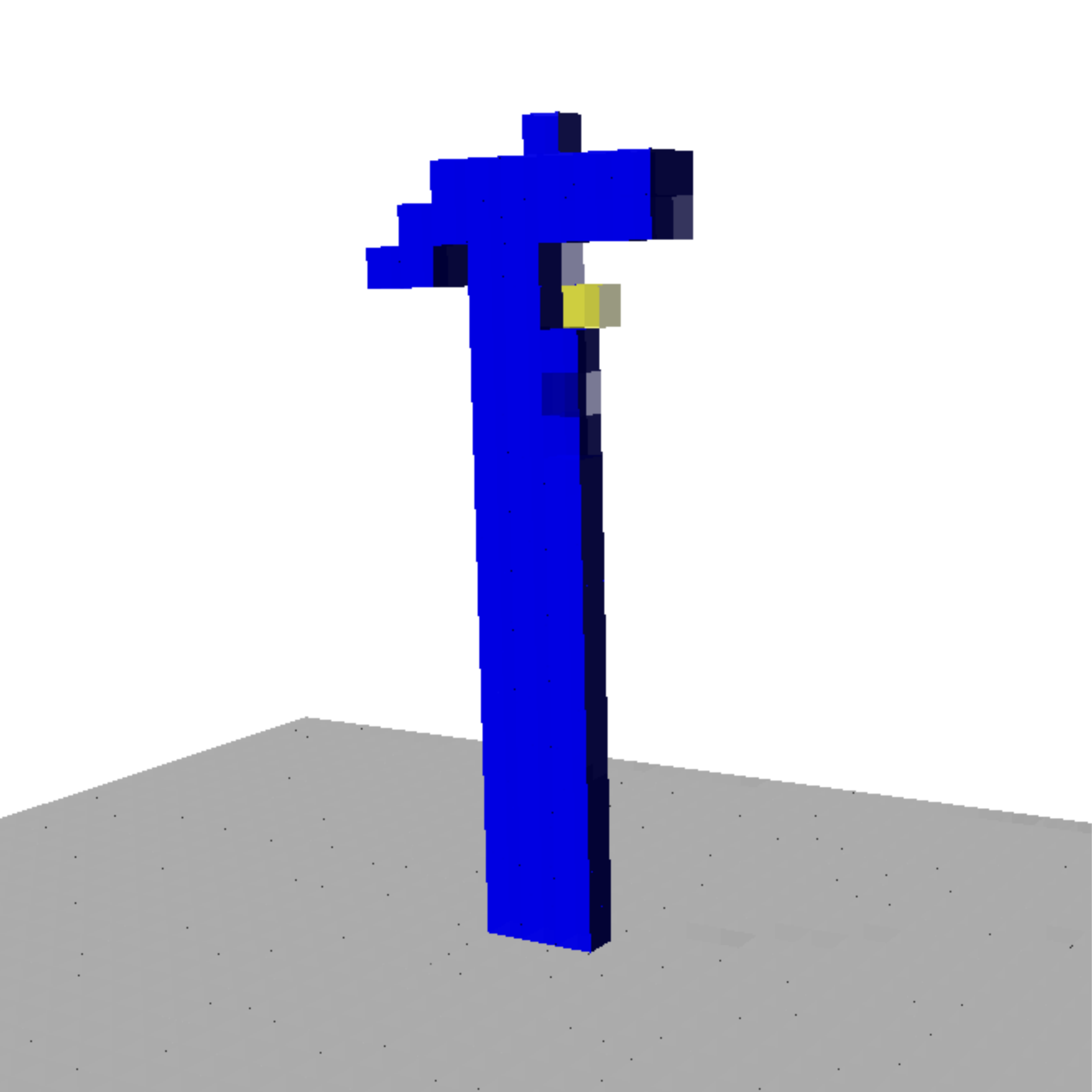}
			\caption{\rtxt{Hammer}}
			\label{fig:hammer}
		\end{subfigure}
		\caption{Probabilistic grids after reconstruction.}
		\label{fig:reconstred_objects}
	\end{figure*}
	
		Table \ref{tab:coverage} presents the percentage of object reconstruction and Table \ref{tab:timem} shows the processing time needed to compute the next-best-view. the method proposed in this work is called Regression, the one proposed in \cite{mendoza2018paper} is labeled as Classif. and the one proposed in \cite{kriegel2012next} based on information gain is labeled  Inf. Gain. \rtxt{Even though less coverage is achieved with the proposed method compared to exhaustive search methods, the resulting models can be good enough for several tasks in which a fast decision is required based on the constructed model. Nonetheless, this can be alleviated by adding a focused exhaustive search stage already available in the literature.}

		\btxt{We underline that the data reported in Table \ref{tab:timem}  called processing times, are the times that the network needs to determine the next-best-view, and not the time they need to learn from examples, that is, the time the network needs to make the inference.} \btxt{We would also like to point out that the reported processing time for regression corresponds to architecture NBV-net 4-5, which is the one with the most layers among the tested regression architectures, and the one that takes the longest to compute the NBV. Therefore, that reported time serves as an upper bound, namely, all the tested regression architectures are able to perform the inference with a frequency of at least 3 Hz.}

\begin{figure}[tb]
		\centering
		\includegraphics[width=\linewidth]{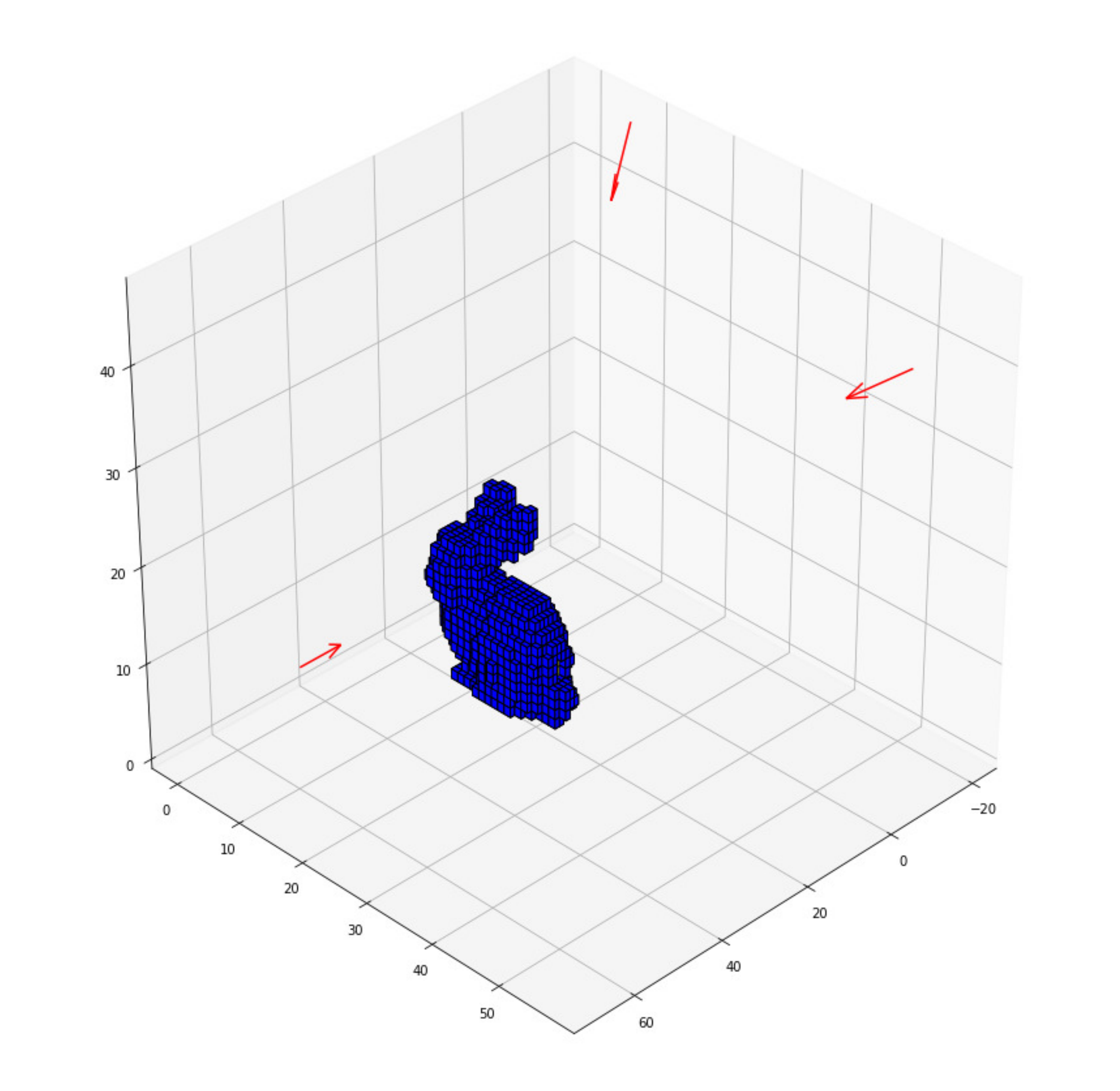}
		\caption{First three sensing locations during the reconstruction of the Bunny object. The arrow in front of the Bunny is the initial view. The remaining arrows were predicted by the Regression method.}
		\label{fig:first_scans}
	\end{figure}

	We can observe that the method proposed in this work, gets a larger percentage of object reconstruction for \rtxt{ten of the thirteen} objects compared with the Classif. method. In consequence, the Regression method improves the coverage reached by the previous Classification method. On the other hand, the Inf. Gain method achieves a larger percentage of object reconstruction in \rtxt{eleven of the thirteen} objects w.r.t. the other two methods. As a result, the exhaustive search provided by Inf. Gain. reaches the highest coverage. However, the Inf. Gain method is two orders of magnitude slower than the one proposed in this paper.
	Thus, the main drawback of the Inf. Gain. approach is the large processing time that the method needs to compute the next-best-view. One can also notice that the faster method is the Classification method, which is an order or magnitude faster than the one proposed in this work. It only requires around 10 milliseconds to compute the next-best-view.
	Nonetheless, the main limitation of the Classification approach is that the fixed number of sensor views could lead to an incomplete model. \rtxt{Also note from the hammer statistics that thin objects are particularly hard to reconstruct for the three methods due to the resolution of the probabilistic grid.} 
	Fig. \ref{fig:reconstred_objects} displays the occupancy grids after reconstruction using the Regression method. Fig. \ref{fig:first_scans} shows an example of the first three sensing views for the Bunny object.

	%\begin{figure}[tb]
	%	\centering
	%	\includegraphics[width=0.7\linewidth]{s7}
	%	\caption{Reconstructed model for the Sarcophagus after 7 scans. Even thought that great part of the surface has been observed, hidden surfaces inside the sarcophagus remain undiscovered.}
	%	\label{fig:sarcho_reconstructed}
	%\end{figure}

	\subsection{Discussion}

	We have found that the proposed method is capable of reconstructing the majority of the object surfaces despite the object shape. The architecture that has the largest number of convolutional and fully connected layers (NBV-net 4-5), was the one with the best result, that is, largest final reconstruction percentage. The main advantage of the proposed method is its rapid response in running time (0.3 seconds). Compared with traditional next-best-view methods, the proposed one eliminates the expensive ray tracing step required to compute several information metrics. Even though training the network might be a time consuming task, it is performed only once and offline. One disadvantage could be that in some cases, it does not reach the highest coverage obtained by search-based methods, for those cases, the current method could be complemented by including a local search or a surface filling.
	
	About the comparison between the method proposed in this work with other two methods. One is a classification-based method \cite{mendoza2018paper} and the other is a exhaustive search using an information gain evaluation \cite{kriegel2012next}. The method based on exhaustive search and information gain gets a larger percentage of object reconstruction, but it is slow. In contrast, the method based on classification is very fast but it gets poor results in terms of percentage of object reconstruction. Based on the data in  Tables \ref{tab:coverage} and \ref{tab:timem}, we conclude that the method proposed in this work gets a good tradeoff between percentage of object reconstruction and  the processing time needed to compute the next-best-view. It overcomes the  method in \cite{mendoza2018paper} in terms of  percentage of object reconstruction, and it needs a reasonable processing time to compute the next-best-view, it only takes less than a half of a second to get it.
	
	With respect to the network training, we believe that the current approach can be improved with a new loss function that does not consider a single NBV as the ground truth. Because in several cases there is more than one good view, and they are separated in distance. However, this is not a trivial task. In addition, for future datasets it will be good to include more object shapes, leaving for the validation set different shapes (not included in the training set) as well as different reconstruction states.

	\section{Conclusions}
	\label{sec:conclusions}
	
	We have presented a deep learning based approach for next-best-view regression. In this approach, we are addressing the next-best-view prediction in a continuous space. The proposed network architecture is designed for the particular problem and it has been trained and validated. Our experiments have shown that the proposed method generalizes well to object shapes that have not been seen by the network during training nor validation. The fast response of the proposed method is one of its advantages given that it eliminates the expensive ray tracing required by state of the art methods. We have presented a comparison  between the method proposed in this work, with other two related approaches.
	We can conclude that the method proposed in this work gets a good trade-off between percentage of object reconstruction and  the processing time needed to compute the next-best-view. For future research, we will study new loss functions as well as applications to the reconstruction of large scale buildings. \btxt{Finally, it is planned to continue expanding the training and validation datasets including additional objects.}

	%\begin{acknowledgements}
	%\end{acknowledgements}


	\bibliographystyle{spmpsci}      % mathematics and physical sciences
	\bibliography{nbv_regression}

\end{document}